\documentclass[dviepsmx]{interact}
\usepackage[subrefformat=parens]{subcaption}

\usepackage[numbers,sort&compress]{natbib}
\bibpunct[, ]{[}{]}{,}{n}{,}{,}
\renewcommand\bibfont{\fontsize{10}{12}\selectfont}
\makeatletter
\def\NAT@def@citea{\def\@citea{\NAT@separator}}
\makeatother
\usepackage{multirow}
\usepackage{siunitx}
\usepackage{url}
\theoremstyle{plain}

\theoremstyle{definition}

\theoremstyle{remark}

\begin{document}

\articletype{FULL PAPER}

\title{Development of Implicit-Explicit Control Based Amphibious Centipede-Type Robot and Evaluation of its Mobile Performance}

\author{
\name{Yusuke Tsunoda\textsuperscript{a}\thanks{CONTACT Yusuke Tsunoda Email: tsunoda@eng.u-hyogo.ac.jp}, Seiya Yamamoto\textsuperscript{b}, Kazuki Ito\textsuperscript{b}, Runze Xiao\textsuperscript{c}, Keisuke Naniwa\textsuperscript{d}, Koichi Osuka\textsuperscript{e}}
\affil{\textsuperscript{a}Graduate School of Engineering, University of Hyogo, Hyogo, Japan; \textsuperscript{b}Graduate School of Engineering, The University of Osaka, Osaka, Japan;\textsuperscript{c}Graduate School of Precision Engineering, The University of Tokyo, Tokyo, Japan;\textsuperscript{d} Engineering, Hokkaido University of Science, Hokkaido, Japan;\textsuperscript{e}Department of Faculty of Robotics and Design, Osaka Institute of Technology, Osaka, Japan}
}

\maketitle

\begin{abstract}
Multi-legged mobile robots possess high mobility performance in rough terrain environments, stemming from their high postural stability, joint flexibility, and the redundancy provided by multiple legs.
In prior research on navigating between different environments such as land and water, the primary strategy employed involves switching to a controller that generates an appropriate gait for the new environment upon entering it.
However, designing appropriate gaits for each complex and diverse environment and accurately determining controller switching for each environment is challenging.
Therefore, this research develops a centipede-type mobile robot that navigates both aquatic and terrestrial environments with a simple, unified control scheme, based on the implicit-explicit control philosophy and by ingeniously designing the robot's body structure.
In this research, we developed the robot featuring flexible joints and left and right legs on each body segment and focused on the leg structure which has extensive contact with the environment. 
This paper evaluates the locomotion performance on land and water using the three developed leg structures, using the robot's leg slip rate and actuator energy consumption as evaluation metrics.
The experimental results confirmed the existence of an appropriate leg structure capable of navigating both aquatic and terrestrial environments under identical control.
\end{abstract}


\section{Introduction}
Multi-legged mobile robots, which move by skillfully manipulating multiple legs like centipedes and millipedes, have long been studied by researchers worldwide.
Key features of these robots include: 1. Stable body support provided by numerous legs, 2. Ability to obtain substantial reaction forces from the surrounding environment, and 3. High fault tolerance, allowing the robot to maintain mobility even if several legs fail.
Consequently, these robots are anticipated as terrain-traversing robots capable of navigating complex, uneven environments.
Prior research on multi-legged mobile robots is extensive and diverse, covering topics such as inter-leg coordination control\cite{aoi2016advantage,chong2022general,yasui2024decentralized}, analysis of the impact of body undulation on locomotion\cite{hoffman2011passive}, demonstration of rough terrain traversal\cite{koh2010centipede,kashiwada2012proposal,ozkan2020systematic,Sfakiotakis2009UndulatoryAP}, navigation\cite{Runze_Xiao2022}, rough terrain exploration\cite{xiao2023unfavorable}, object transport\cite{chong2023multilegged,ozkan2021self}, self-assembling robots through combination and separation\cite{app14062331,ozkan2021self}, and analysis of actual arthropods and their application to robotics\cite{yasui2019decoding,diaz2023active,mingchinda2023leg}.

Among these, the development of multi-legged mobile robots capable of traversing different environments—land and water (underwater)—is a particularly important challenge.
Previous research has reported several examples of such robot development.
For example, RHex is a representative six-legged mobile robot capable of traversing land, gravel environments, mud, and water. It can navigate diverse terrains using a simple leg mechanism and gait control\cite{Rhex}.
This robot rotates its arc-shaped legs at different phases for land movement and drives its fin-shaped legs in phase for swimming.
Ijspeert et al. reported a biomimetic robot inspired by the amphibian salamander\cite{salamander}.
The developed robot generates gaits based on phase oscillators driven by central pattern generators (CPGs) and switches gaits between terrestrial and aquatic environments, enabling smooth transitions between land and water.
Baines et al. drew inspiration from the adaptive locomotion of turtles in aquatic and terrestrial environments, proposing a variable-stiffness fin mechanism and corresponding walking, crawling, and swimming motions optimized for each environment\cite{baines2022multi}.
These robots require switching between appropriate body configurations, gait generation, and control for movement in terrestrial and aquatic environments.
However, in the complex and diverse conditions of real-world natural environments, it is difficult to pre-design appropriate gaits and body configurations for each environment.
Furthermore, even if this were possible, the system would need to detect environmental changes from sensory input and switch controllers or body configurations, potentially leading to increased complexity.
In contrast, if navigation between both environments could be achieved using a unified body and a single, simple control scheme, the computational resources required by the robot could be significantly reduced.

\begin{figure}[t]
  \centering
  \includegraphics[width=0.7\columnwidth]{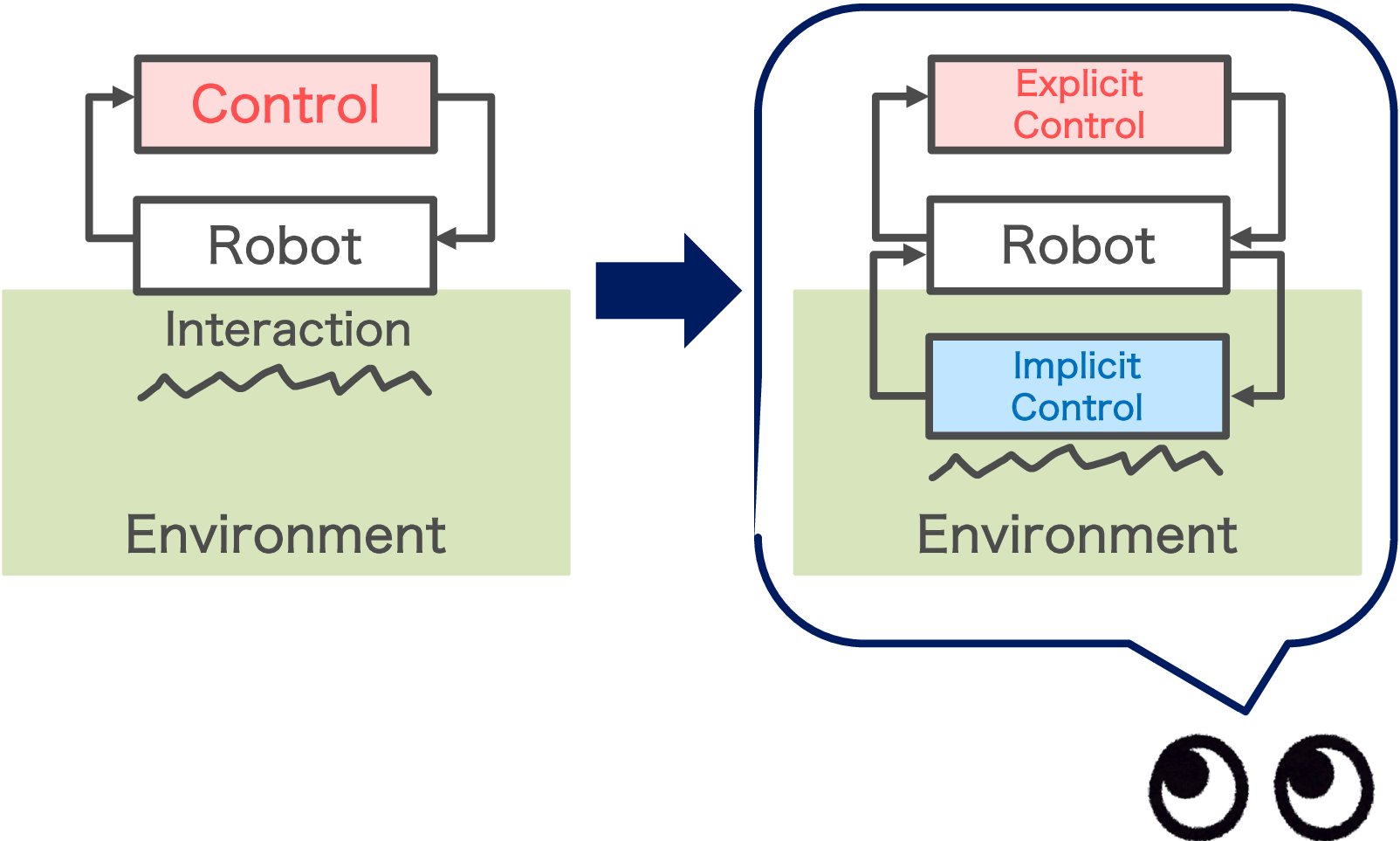}
  \caption{Strucuture of Explicit-Implicit Control}
  \label{fig:implicit}
\end{figure}

\begin{figure}[t]
  \begin{minipage}[t]{0.5\columnwidth}
   \centering
   \includegraphics[width=\columnwidth]{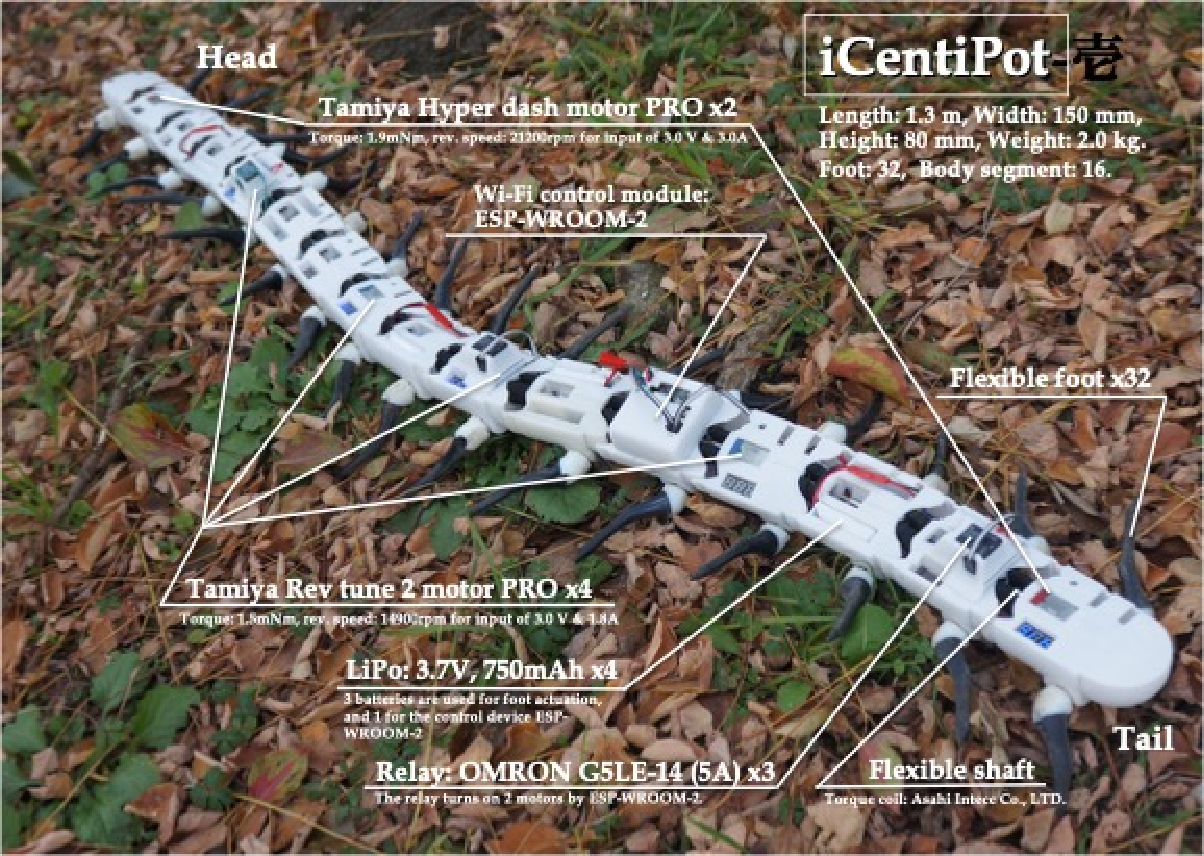}
   \subcaption{i-CentiPot\cite{kinugasa2017development,osuka2019centipede}}
   \label{fig:i_centipot}
  \end{minipage}
  \begin{minipage}[t]{0.5\columnwidth}
   \centering
   \includegraphics[width=\columnwidth]{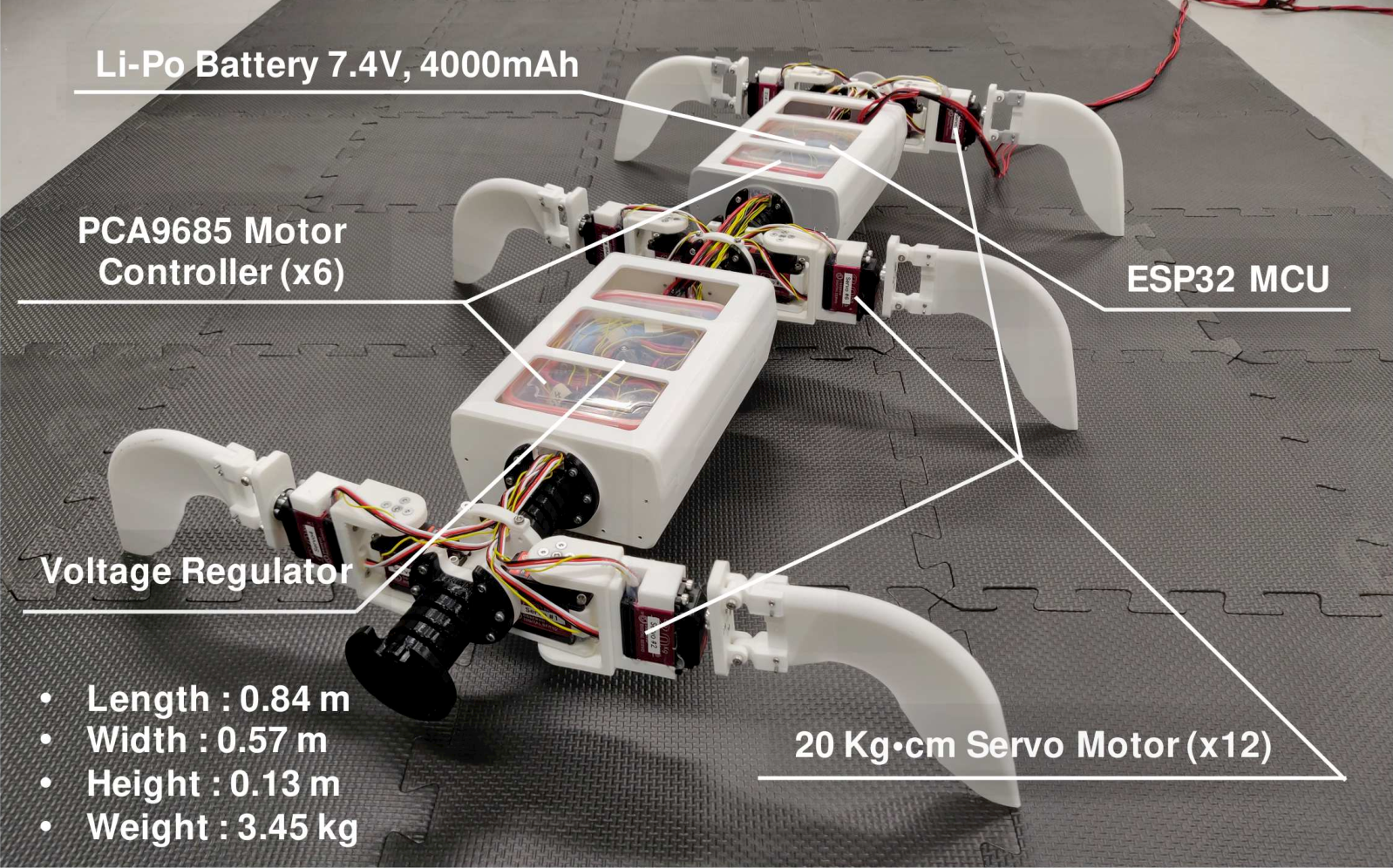}
   \subcaption{i-AMMR\cite{Tieng18042025}}
   \label{fig:i_ammr}
  \end{minipage}
  \caption{Implicit-Explicit control based multi-legged mobile robots in the previous researches}
  \label{fig:previous_robots}
\end{figure}

\begin{figure}[t]
  \centering
  \includegraphics[width=0.6\columnwidth]{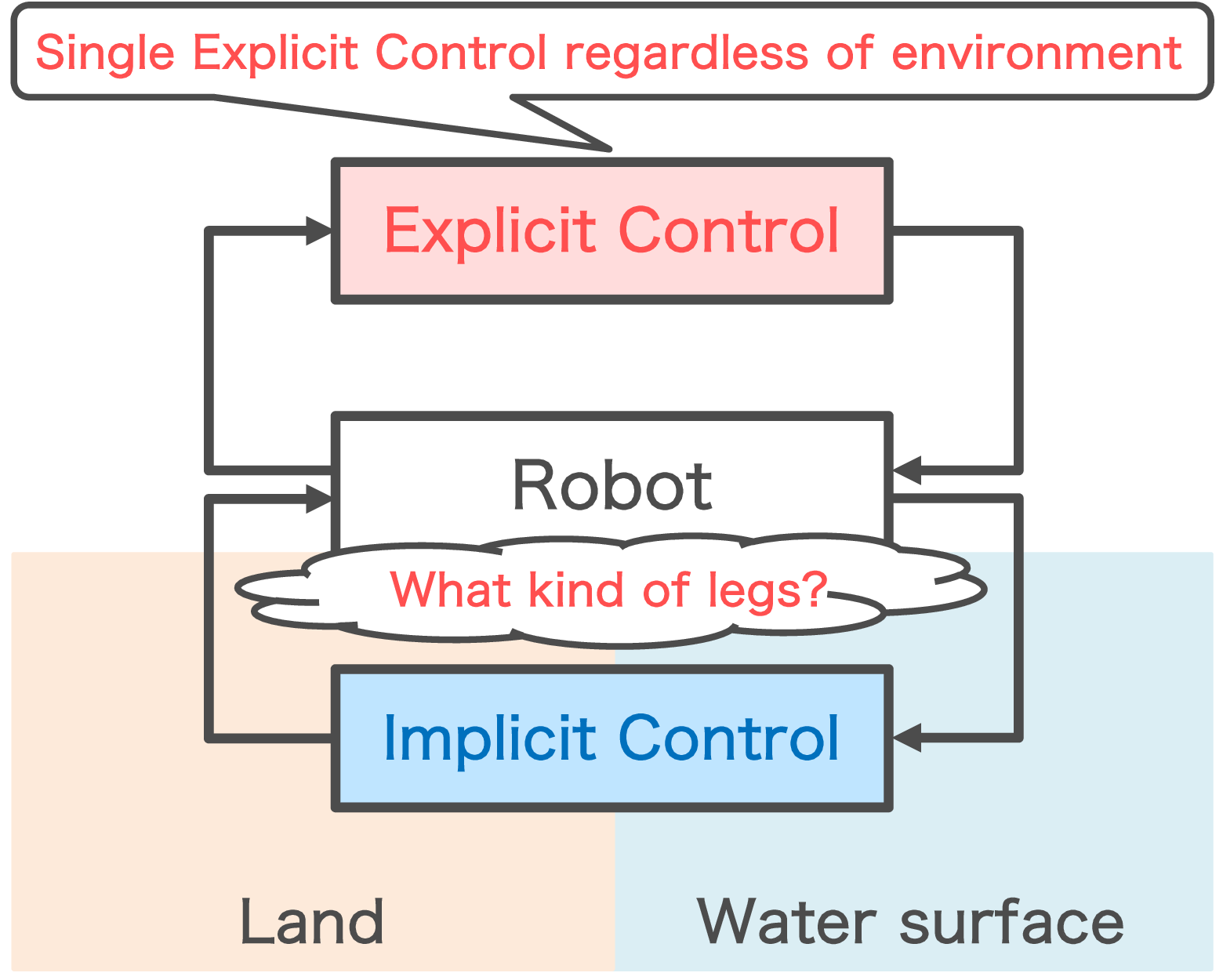}
  \caption{Approach of this research}
  \label{fig:approach}
\end{figure}

To develop such robots, we draw inspiration from two concepts: implicit-explicit control \cite{osuka2010dual} and open design \cite{open}.
Implicit-Explicit Control, as shown in Fig.~\ref{fig:implicit}, is a concept that recommends designing the body structure and dynamics so that the mechanical interaction between the environment and the robot is not treated as a disturbance, but rather becomes an effective control (implicit control).
Open design, meanwhile, provides design guidelines for robots navigating environments that change unpredictably and constantly, such as natural environments—referred to as unspecified environments.
Unlike the constrained environments traditionally assumed, unspecified environments have open boundary conditions. 
This means it is impossible to design optimal control or a body within them.
Therefore, we need to obtain an optimal solution for robot design in some sense.
We argue that we must derive suboptimal solutions within the framework of implicit control and develop robots that perform reasonably well in various environments.
In prior research \cite{kinugasa2017development,osuka2019centipede}, the multi-legged mobile robot “i-CentiPot” developed by Osuka et al., as shown in Fig.~\ref{fig:i_centipot} demonstrates results that strongly support these ideas.
This robot consists of 8 segments and 16 legs, rotating each leg by transmitting torque from multiple DC motors through a single torque tube.
The i-CentiPot features flexible passive joints between segments and legs formed from flexible materials.
Consequently, despite each leg rotating at a single angular velocity, the robot can move while adapting to environmental undulations and utilizing interactions with its surroundings through simple control.
Based on this i-CentiPot, we aim to develop an amphibious centipede-like robot capable of moving in both aquatic and terrestrial environments with simple control.
As a precursor, we developed the Implicit amphibious multi-legged mobile robot (i-AMMR) in \cite{Tieng18042025}, a six-legged mobile robot based on implicit-explicit control, as shown in Fig.~\ref{fig:i_ammr}. 
We experimentally explored the inter-segmental joint stiffness for optimal terrestrial locomotion efficiency and the control patterns for each leg. 
However, with only six legs, experimental and performance evaluations as a centipede-type robot were insufficient.

In this study, we further develop the robot described in \cite{Tieng18042025} by refining its body structure to create a centipede-like mobile robot capable of navigating both aquatic and terrestrial environments with a single control system.
We adopt an approach based on the implicit-explicit control structure shown in Fig.~\ref{fig:approach}.
As shown in Fig.~\ref{fig:approach}, we focus on the leg structure where interactions between the environment and the robot's body primarily occur, and develop a leg structure that enables implicit control of interactions between the robot and the environment in both aquatic and terrestrial environments.
We propose new flexible legs equipped with fins for swimming in water.
We then comprehensively experimentally verified how the presence or absence of fins on the legs, the presence or absence of leg flexibility, and differences in phase difference control between legs affect mobility performance in aquatic and terrestrial environments, evaluating the mobility performance.
This study adopts leg slip rate and actuator energy consumption as evaluation parameters for mobility performance.
The slip rate is an indicator that accounts for the differing rotational radii of each leg structure proposed in this paper, evaluating mobility performance based on the contribution rate of each leg's circumferential velocity to the robot's overall travel speed.
Furthermore, by comparing the energy consumption of the robot's actuators across each parameter set, we examine which parameter set dominantly utilizes the interaction between the environment and the body for locomotion.

The structure of this paper is as follows.
Section \ref{sec:robot} provides an overview of the amphibious centipede robot developed in this paper.
Section \ref{sec:expe} presents evaluation experiments assessing the robot's mobility performance in both land and water environments.
Section \ref{sec:result} presents the experimental results, and Section \ref{sec:discussion} discusses the mobility performance based on evaluation metrics.
Finally, Section \ref{sec:conc} concludes this paper.

\begin{figure}[t]
  \centering
  \includegraphics[width=0.7\columnwidth]{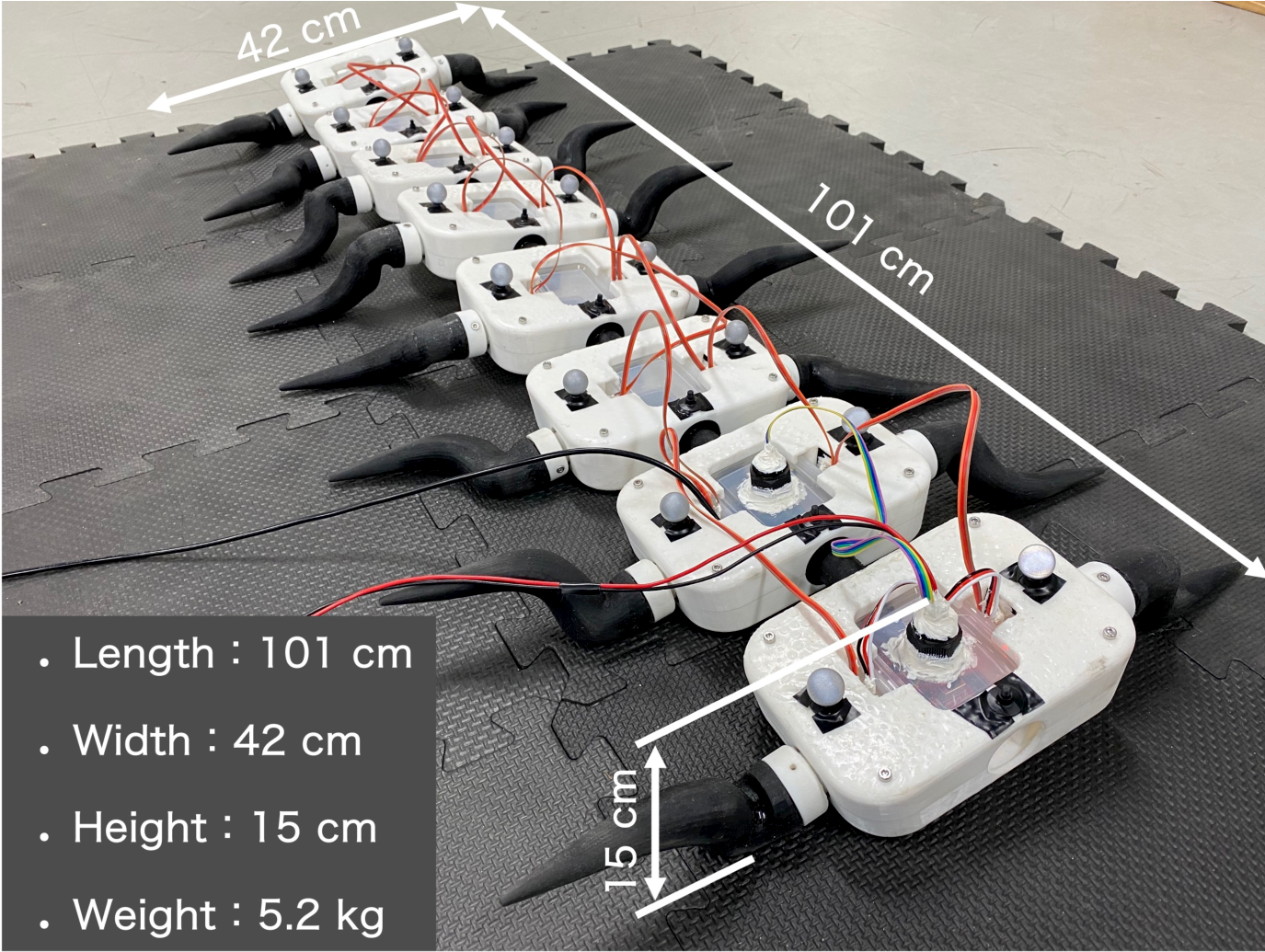}
  \caption{i-CentiPot-Amphibian}
  \label{fig:i-CPA}
\end{figure}

\begin{table}[t]
  \caption{Specification of i-CentiPot-Amphibian}
  \label{tab:size}
  \centering
  \begin{tabular}{cc}\hline
  Specifications & Values \\\hline\hline
  Length & 101~\textrm{cm} \\
  Width & 42~\textrm{cm} \\
  Height & 15~\textrm{cm} \\
  Weight & 5.2~\textrm{kg} \\ \hline
  \end{tabular}
\end{table}

\begin{figure}[t]
  \centering
  \includegraphics[width=0.7\columnwidth]{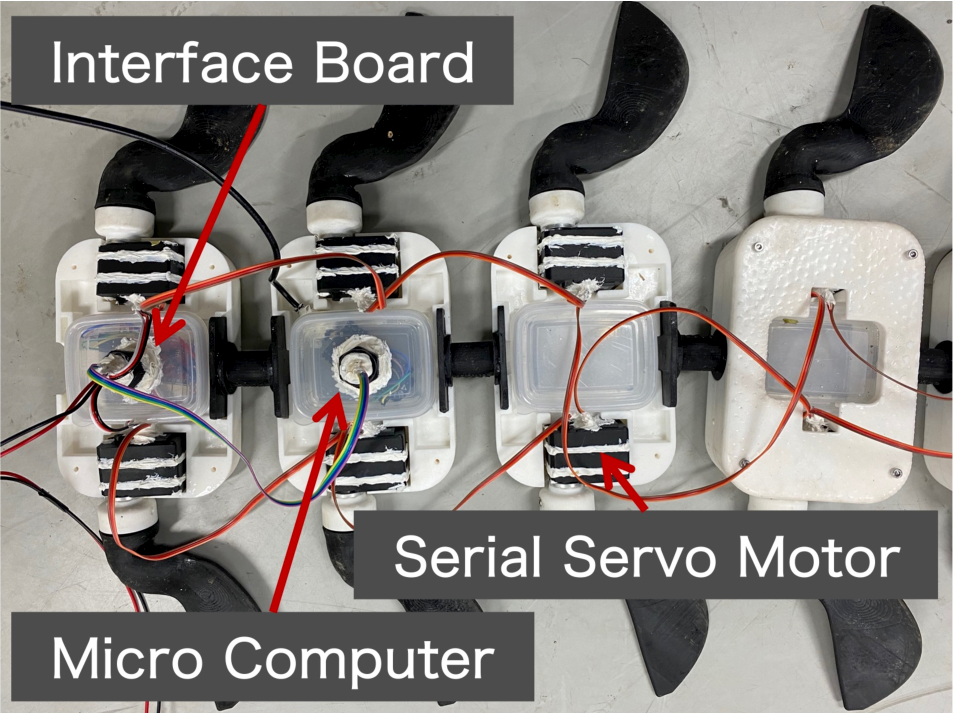}
  \caption{Inside parts of i-CentiPot-Amphibian}
  \label{fig:inner}
\end{figure}

\begin{figure}[t]
  \centering
  \includegraphics[width=0.7\columnwidth]{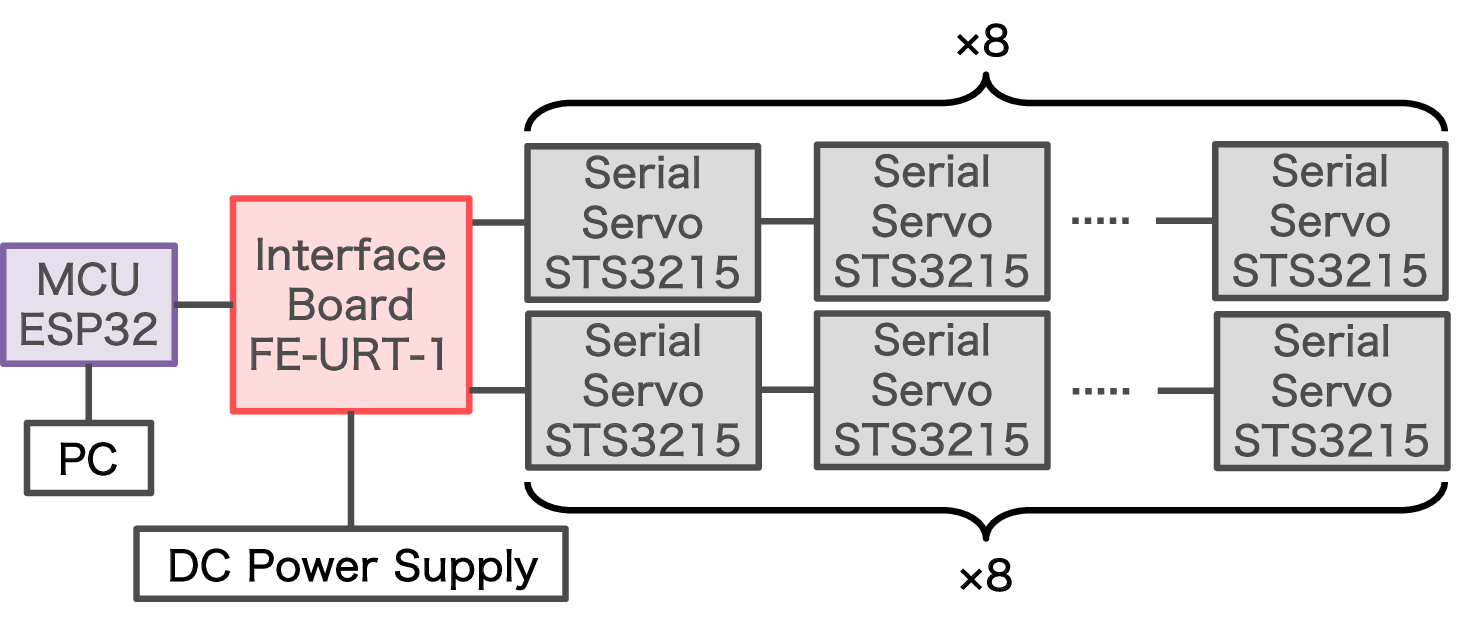}
  \caption{System of i-CentiPot-Amphibian}
  \label{fig:kairo}
\end{figure}

\begin{table}[t]
  \caption{Mechanical and electrical  of i-CentiPot-Amphibian}
  \label{tab:parts}
  \centering
  \begin{tabular}{lll}\hline
  Part & Name & Quantity\\\hline \hline
  Serial servo motor & 7.4V 19kg$\cdot$cm magnetic encoding servo STS3215 & 16\\ 
  Interface board & FE-URT-1 USB to TTL 485 bus programmer & 1\\
  Micro computer & ESP32 DevKitC V4 ESP-WROOM-32 ESP-32 WiFi & 1\\\hline
  \end{tabular}
\end{table}

\section{Development of the amphibious centipede-type robot “i-CentiPot-Amphibian”}
\label{sec:robot}
This research aims to develop a multi-legged robot capable of flexible movement through both land and water environments using a single, simple active control system independent of the environment.
As mentioned in the introduction, we developed the amphibious centipede-type mobile robot “i-CentaPot-Amphibian (i-CPA)” by referencing the mechanism and control of the i-CentaPot robot.
This chapter describes the conceptual design of i-CPA, the resulting body design, and the control system design for each leg.

\subsection{Conceptual design}
\label{sec:gainen}
i-CentiPot moves on land by conforming to environmental undulations through its body's structural flexibility.
However, movement on water surfaces is not anticipated, and its body structure is expected to be unsuitable for propelling itself through water.
The robot developed in this study must be driven by a single control system like i-CentiPot, while possessing both a flexible body for land movement and a body structure capable of swimming on water.
This research focuses on the leg structure, where the primary interaction between the environment and the robot's body occurs in both terrestrial and aquatic environments.
Therefore, we decided to develop legs that can generate sufficient ground reaction force for terrestrial movement and possess flexible fins to effectively propel through water.

\subsection{Overview of the developed robot}
\label{sec:diagram}
Fig.~\ref{fig:i-CPA} shows the appearance of the developed i-CPA, and Table~\ref{tab:size} lists its various dimensions.
The i-CPA is an 8-segmented, 16-legged amphibious mobile robot featuring flexible flipper-like legs and joints.
The robot body and joints were fabricated using a stereolithography printer.
Specifically, the joints use Rubber-like Resin F80 Elastic 3D Printer Resin, providing flexibility.
Fig.~\ref{fig:inner} shows the internal structure of each body segment.
Each segment symmetrically houses two infinitely rotatable serial servo motors, with legs attached to the motors' output shafts.
To achieve waterproofing, the motors are sealed with silicone, and other electronic components are housed inside a Tupperware container.
The Tupperware container also serves as a float to keep the robot afloat in water.

Fig.~\ref{fig:kairo} shows a schematic diagram of the i-CPA system, and Table~\ref{tab:parts} lists the mechanical and electronic components used.
Sixteen serial servo motors are connected in series to the control board, which is connected to the microcontroller.
Power for driving the motors is supplied by a DC stabilized power supply.
Sensors are built into the motors, enabling acquisition of information on motor phase, rotational speed, input voltage, and current consumption.

\begin{figure}[t]
  \begin{minipage}[t]{0.33\columnwidth}
    \centering
    \includegraphics[width=\columnwidth]{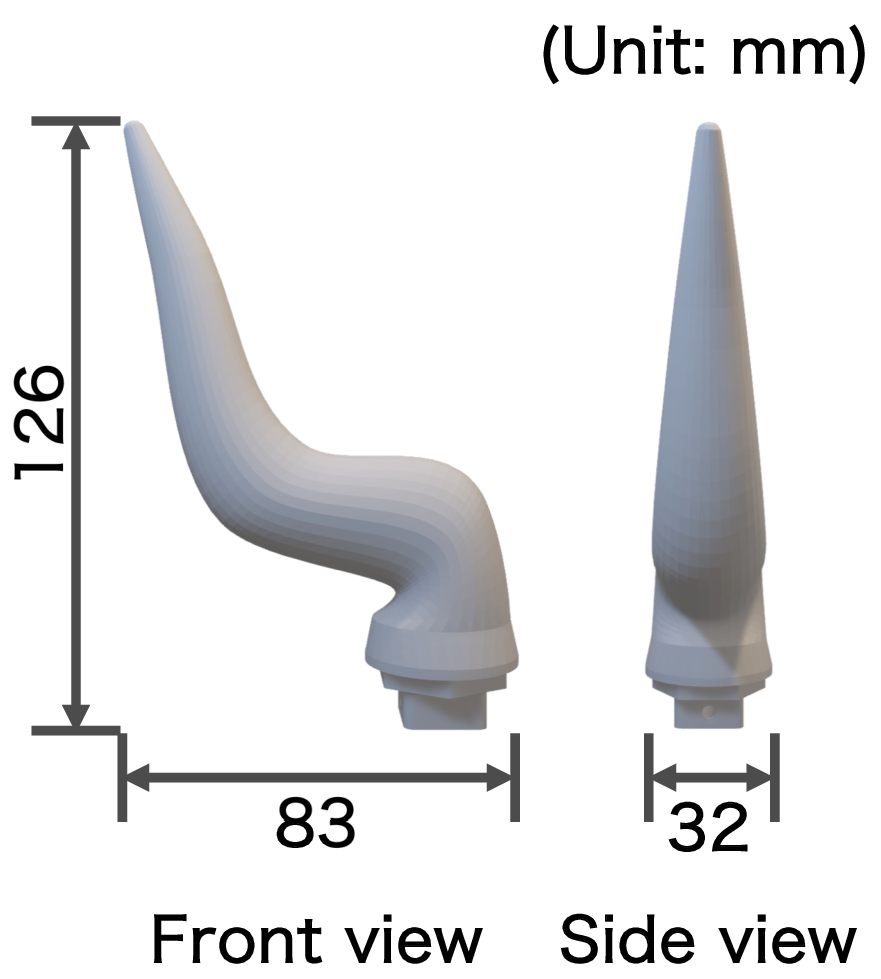}
    \subcaption{The leg of Normal type}
    \label{fig:Normal}
  \end{minipage}
  \begin{minipage}[t]{0.33\columnwidth}
    \centering
    \includegraphics[width=\columnwidth]{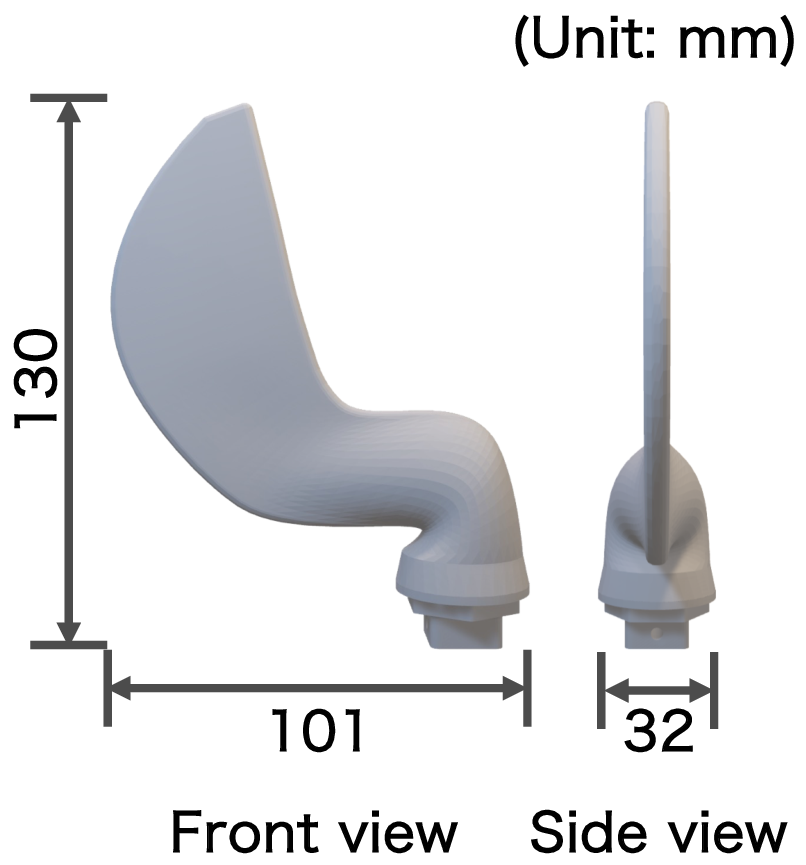}
    \subcaption{The leg of Fin type}
    \label{fig:Fin}
  \end{minipage}
  \begin{minipage}[t]{0.33\columnwidth}
    \centering
    \includegraphics[width=\columnwidth]{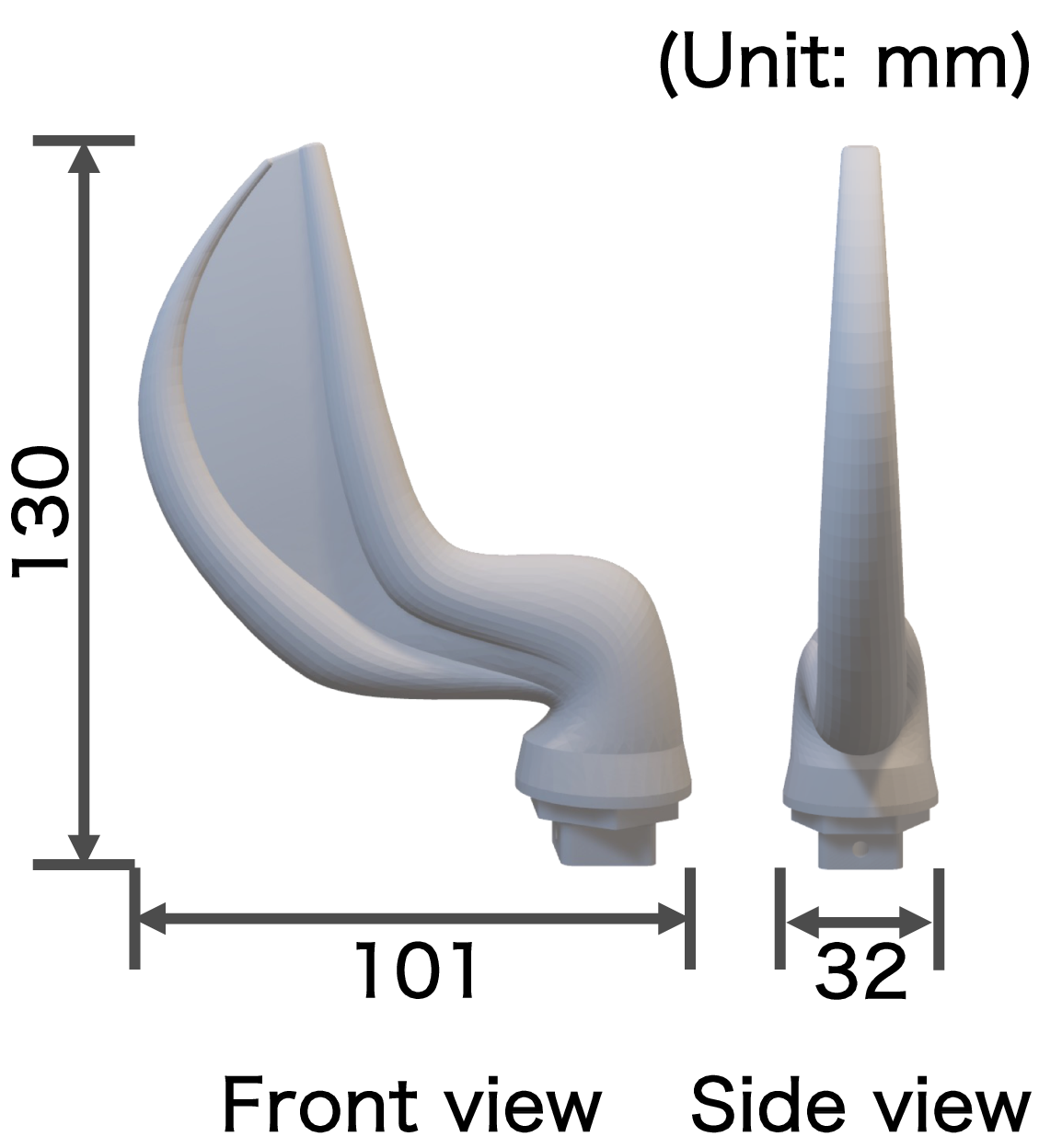}
    \subcaption{The leg of Web type}
    \label{fig:Web}
  \end{minipage}
  \caption{Leg types of i-CentiPot-Amphibian}
  \label{fig:leg}
\end{figure}

\subsection{Fin leg design}
\label{sec:fin}
The three designed leg shapes and their main dimensions are shown in Fig.~\ref{fig:leg}.
All legs were fabricated using a stereolithography printer, employing rubber-like resin F80 Elastic 3D Printer Resin to achieve flexibility.
In this study, we focused on leg shape and flexibility in the design, considering that for a robot to effectively utilize interaction with its environment, it is crucial for the legs to deform in response to environmental undulations and increase their contact area.
The Normal type leg was designed based on the leg shape of the i-CentiPot.
The stiffness of the legs is designed to be rigid at the base to lift the body on land and progressively flexible toward the tip to conform to the environment.
Therefore, by making the base thicker and tapering toward the tip, legs that are rigid at the base and progressively softer toward the tip are achieved.
In contrast, the Fin type features fin-like legs designed to increase the area for water displacement to maximize thrust generated by paddling.
The stiffness of the legs is designed to be rigid at the base to lift the body on land, while the fin portion is made thin to be flexible and adapt to the environment.
The Web type combines features of both the Normal and Fin types, featuring a webbed shape.
The thinness of the water-propelling section matches the Fin type, while the section itself is shaped like the Normal type's legs to function as a webbed paddle. The stiffness of the legs is achieved by making the base rigid, while maintaining a degree of flexibility from the base to the tip, though it is stiffer than the Fin type.
Furthermore, when moving on water, if the legs scoop water in the direction of travel, it hinders the i-CPA's progress.
Therefore, the design of each leg is intended so that when scooping water in the direction of travel, the leg emerges from the water as much as possible.

\subsection{Fin leg control}
\label{sec:control}
To confirm that the flexible fin-like legs enable effective mechanical interaction between the robot and its environment for locomotion in both aquatic and terrestrial environments, this study aims to keep the explicit control (positive control) applied to the robot as simple as possible.
The bus servo motors connected to each leg of the i-CPA can independently control the rotational angular velocity of the legs.
To achieve traversal with the simplest possible control, each leg rotates at the same angular velocity in both terrestrial and aquatic environments.
The control applied to the robot consists of all legs rotating at a constant angular velocity, with two types of phase difference between left and right legs: in-phase and out-of-phase.
Details of the control used in the experiments are described in Sec.~\ref{sec:condition}.

\begin{figure}[t]
  \begin{minipage}[t]{0.5\columnwidth}
   \centering
   \includegraphics[width=\columnwidth]{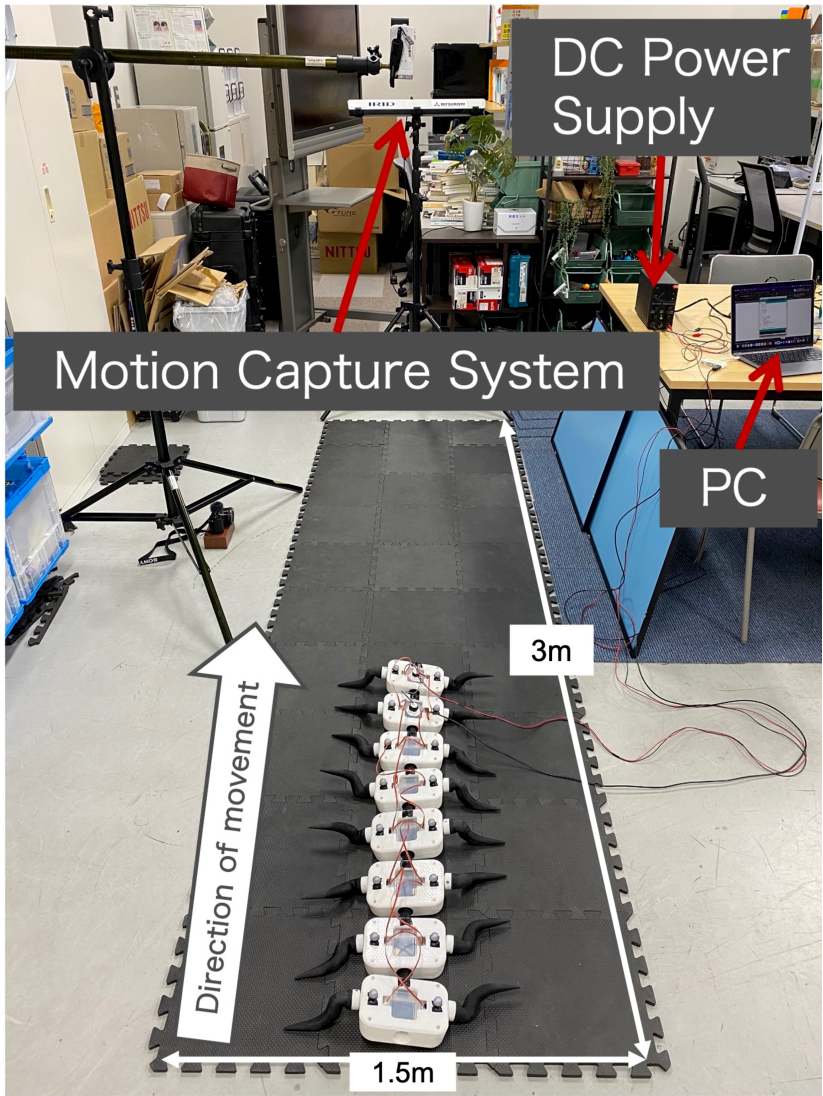}
   \subcaption{Land environment}
   \label{fig:land environment}
  \end{minipage}
  \begin{minipage}[t]{0.5\columnwidth}
   \centering
   \includegraphics[width=\columnwidth]{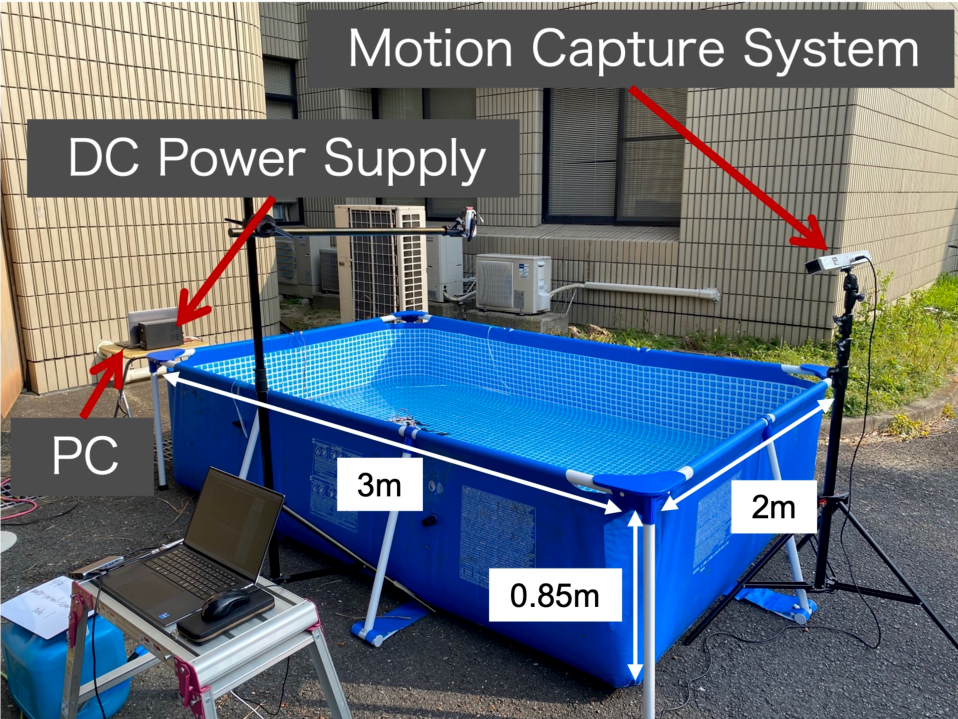}
   \subcaption{Water surface environment}
   \label{fig:water surface environment}
  \end{minipage}
  \caption{The experiment environments}
  \label{fig:area}
\end{figure}

\begin{figure}[t]
  \centering
  \includegraphics[width=0.3\columnwidth,angle=-90]{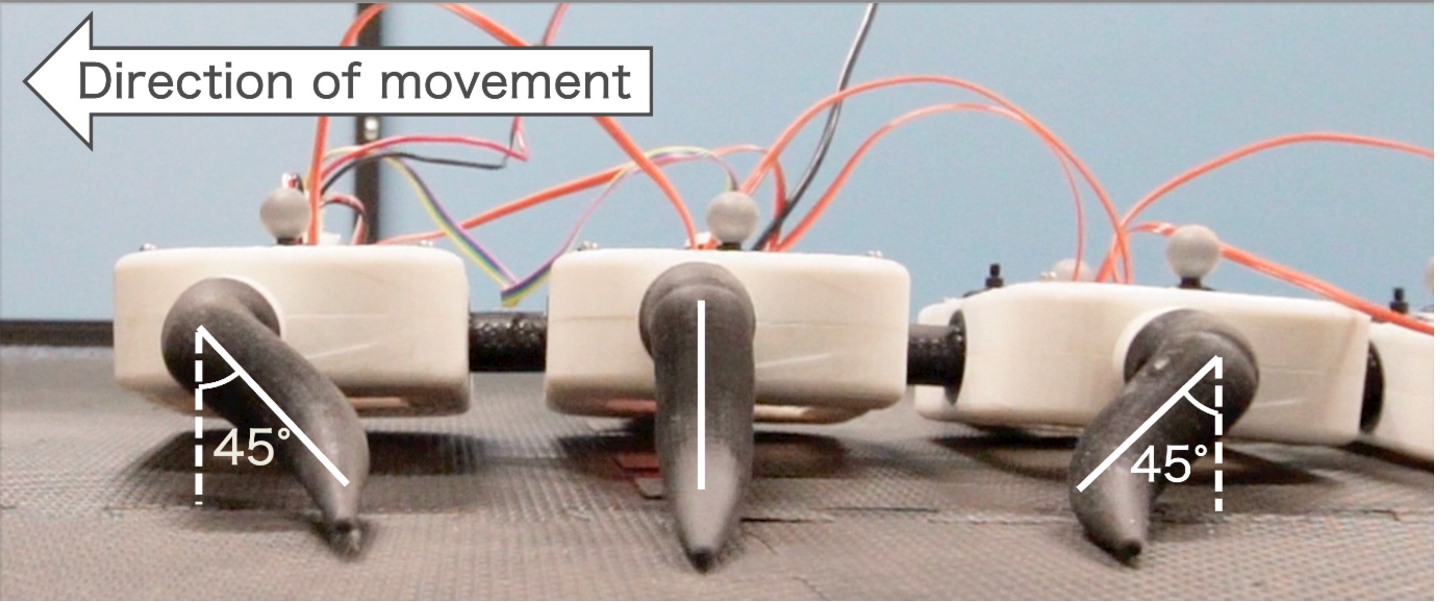}
  \caption{Phase between consecutive legs which generates retrograde wave}
  \label{fig:isou}
\end{figure}

\begin{figure}[t]
  \centering
  \includegraphics[width=0.7\columnwidth]{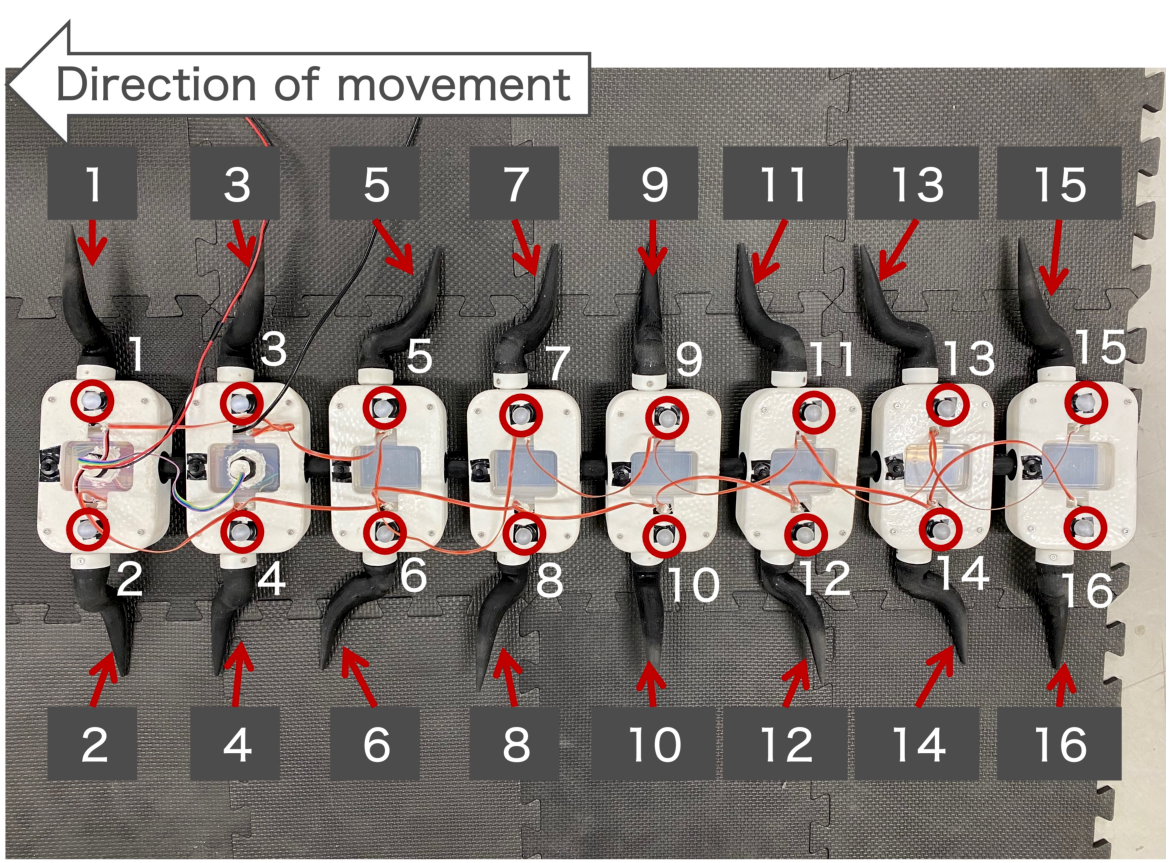}
  \caption{Number of leg and motion capture marker}
  \label{fig:number}
\end{figure}

\section{Experiments for evaluating mobility performance}
\label{sec:expe}
The purpose of this experiment is to investigate which leg shape is best suited for locomotion in both terrestrial and aquatic environments.
To this end, locomotion performance evaluation experiments were conducted in both terrestrial and aquatic environments using three leg types: Normal, Fin, and Web.

\subsection{Experimental environment}
\label{sec:env}
The experimental setup is shown in Fig.~\ref{fig:area}.
Land experiments involved moving the i-CPA over an EVA mat, while water experiments involved moving it over the surface of a pool.
A DC stabilized power supply was used to power the i-CPA.
Additionally, the OptiTrack V120:TRIO motion capture system was used to acquire motion trajectory data for the 16 markers attached to the i-CPA.
Data on the motor phase, rotational speed, input voltage, and current consumption were acquired via sensors built into the serial servo motors.

\subsection{Experimental conditions}
\label{sec:condition}
The experimental conditions are described below.
First, two environments were prepared: land and water.
Three leg shapes were prepared: Normal, Fin, and Web.
Additionally, the phase of the left and right legs was set under two conditions: opposite phase and same phase.
The phase difference between the front and rear legs was set to 45 degrees, but the rotation direction when shifting was the direction in which the leg generates a rearward wave, as shown in Fig.~\ref{fig:isou}.
That is, after a leg contacts the ground, the next rear leg contacts the ground when it rotates 45 degrees.
Combining the above conditions resulted in 2 (environment) × 3 (leg shape) × 2 (left/right leg phase) = 12 conditions. Five experiments were conducted for each condition.

\begin{figure}[t]
  \begin{minipage}[t]{0.5\columnwidth}
   \centering
   \includegraphics[width=0.7\columnwidth]{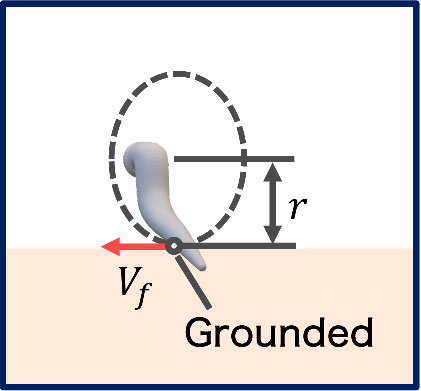}
   \subcaption{Turning radius of leg when moving on land}
   \label{fig:land}
  \end{minipage}
  \begin{minipage}[t]{0.5\columnwidth}
   \centering
   \includegraphics[width=0.7\columnwidth]{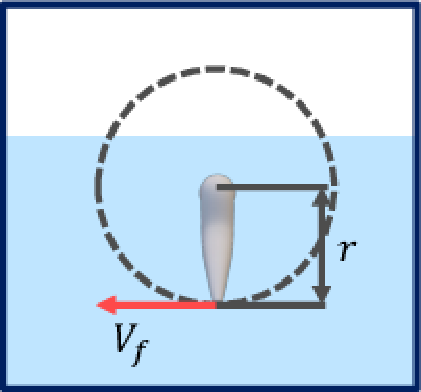}
   \subcaption{Turning adius of leg when moving on water surface}
   \label{fig:water surface}
  \end{minipage}
  \caption{Turning radius of leg when moving on both land and water surface}
  \label{fig:r}
\end{figure}

\begin{table}[t]
  \caption{Turning radius of legs on both land and water surface}
  \label{tab:leg_r}
  \centering
  \begin{tabular}{ccc}\hline
  Type & Land & Water surface\\\hline \hline
  Normal & 50~\si{mm} & 67~\si{mm}\\ 
  Fin & 53~\si{mm} & 86~\si{mm}\\
  Web & 83~\si{mm} & 86~\si{mm}\\\hline
  \end{tabular}
\end{table}

\subsection{Evaluation metrics}
\label{sec:evaluate}
This chapter discusses evaluation metrics for a robot's mobility performance: slip rate and energy consumption.

\subsection{Evaluatuin indicator 1: Slip rate}
One of the simplest performance evaluation metrics for mobile robots is travel speed.
However, since the center of gravity travel speed of the i-CPA also depends on the leg rotation radius, there is an issue where the contribution of leg shape differences to mobility performance becomes unclear.
Therefore, to ignore the influence of the leg rotation radius, we define the slip ratio $\alpha$.
The slip ratio $\alpha$ is an indicator representing the ratio of the robot's travel speed to the leg's circumferential speed, defined by the following equation.

\begin{align}
  \alpha= \frac{V}{V_f}\times100~[\%],
  \label{eq:suberi}
\end{align}

where $V$~[\si{mm/s}] is the time-averaged translational velocity of the i-CPA, and $V_f$~[\si{mm/s}] is the 16-leg average of the time-averaged rotational velocity of each leg.
A larger slip ratio $\alpha$ indicates a greater contribution of the robot's leg rotational velocity to its translational velocity.
As shown in Fig.~\ref{fig:number}, legs and markers are numbered and referred to as Leg~1 to Leg~16 and Marker~1 to Marker~16, respectively.
$V$ is calculated as follows using the motion trajectory data from the $T$ seconds during which the 16 reflective markers attached to the i-CPA were in motion.

\begin{eqnarray}
  V_x&=&\frac{1}{16}\sum_{i=1}^{16}\left(\frac{1}{Tf-1}\sum_{j=1}^{Tf-1}\frac{x_i(t_{j+1})-x_i(t_{j})}{t_{j+1}-t_{j}}\right),\\
  V_y&=&\frac{1}{16}\sum_{i=1}^{16}\left(\frac{1}{Tf-1}\sum_{j=1}^{Tf-1}\frac{y_i(t_{j+1})-y_i(t_{j})}{t_{j+1}-t_{j}}\right),\\
  V&=&\sqrt{V_x^2+V_y^2},
  \label{eq:v}
\end{eqnarray}
where $f$ is the measurement frequency of the motion capture system, $x_i(t_{j})$~[\si{mm}] is the $x$ coordinate of Marker~$i$ ($i=1,\dots,16$) at time $t_{j}=0.008333j$~ [si{s}] in a Cartesian coordinate system with an $xy$ plane parallel to the ground, and $y_i(t_{j})$ [si{mm}] is the $y$ coordinate of Marker $i$ at time $t_{j}$.
In this study, the measurement frequency is $120$ [fps], and the measurement time is $T=7.0$ [s].
Furthermore, $V_f$ is calculated using the following equation.

\begin{align}
  V_f=\frac{1}{16}\sum_{i=1}^{16}2\pi r N_i,
  \label{eq:roundv}
\end{align}
where $r$~[\si{mm}] is the leg rotation radius, and $N_i$~[rps] ($i=1,\dots,16$) is the time-averaged rotational speed of Leg $i$.
For example, when the robot is not moving forward at all, $V=0$~\si{mm/s}, resulting in $\alpha=0$~\%.
Furthermore, when using non-slip tires on the legs, $V=V_f$, and $\alpha=100$
Thus, $\alpha$ expresses how fast the robot can move, independent of the rotational radius.

However, as shown in Fig.~\ref{fig:r}, on land, $r$ is the height of the leg's rotational center above the ground, while in water, $r$ is the maximum radius from the leg's rotational axis.
The values of $r$ for the three leg types are shown in Table~\ref{tab:leg_r}.
The $r$ values for land are the measured heights of the leg's center of rotation above the ground when all legs are in vertical contact with the ground. The $r$ values for water are the measured maximum radii from the leg's rotational axis when no external load is applied to the leg.

\subsection{Evaluatuin indicator 1: Energy consumption}

The second evaluation metric is the total energy consumption of the actuators mounted on each leg of the i-CPA.
The total energy consumption is calculated by integrating the current consumption of the motors mounted on each leg of the i-CPA, as shown in the following equation.

\begin{eqnarray}
  E&=&\frac{1}{16}\sum_{i=1}^{16}\left(\frac{1}{T}\sum_{j=1}^{T} e_i(t_j)\right),\\
  e_i(t_j)&=& A_i(t_j)V_i(t_j),
\end{eqnarray}
where $E$~[\si{J}] denotes the total energy consumption of the actuators mounted on each leg of the i-CPA, $T$ is the experimental time, $e_i(t_j)$~[\si{J}] is the energy consumption of the motors on Leg $i$ at time $t_j$, $A_i(t_j)$~[\si{A}] is the current consumption of the motor of Leg $i$ at time $t_j$, and $V_i(t_j)$~[\si{V}] is the input voltage of the motor of Leg $i$ at time $t_j$.

\begin{table}[t]
  \caption{Experimental results of the velocity $V$ and each leg's velocity $V_f$ in the case that left and right legs are antiphase}
  \label{tab:performance_anti_v}
  \centering
  \begin{tabular}{c|c|c|c|c}\hline
    \multirow{2}{*}{Type}&\multicolumn{2}{c|}{Land}&\multicolumn{2}{c}{Water surface}\\\cline{2-5}
     &$V$ & $V_f$  &$V$ & $V_f$ \\\hline \hline
     Normal&63.4~\si{mm/s}& 153.2~\si{mm/s}  &90.8~\si{mm/s} & 205.4~\si{mm/s} \\
     Fin&85.4~\si{mm/s} & 162.3~\si{mm/s} &120.0~\si{mm/s} & 263.3~\si{mm/s} \\
     Web&78.1~\si{mm/s} & 254.4~\si{mm/s} &88.4~\si{mm/s} & 264.4~\si{mm/s} \\ \hline
  \end{tabular}
\end{table}

\begin{table}[t]
  \caption{Experimental results of the velocity $V$ and each leg's velocity $V_f$ in the case that left and right legs are in-phase}
  \label{tab:performance_in_v}
  \centering
  \begin{tabular}{c|c|c|c|c}\hline
    \multirow{2}{*}{Type}&\multicolumn{2}{c|}{Land}&\multicolumn{2}{c}{Water surface}\\\cline{2-5}
     &$V$ & $V_f$ &$V$ & $V_f$ \\\hline \hline
     Normal&54.0~\si{mm/s} & 153.4~\si{mm/s}  &91.4~\si{mm/s} & 205.2~\si{mm/s} \\
     Fin&65.8~\si{mm/s} & 162.5~\si{mm/s} &112.2~\si{mm/s} & 263.3~\si{mm/s} \\
     Web&82.8~\si{mm/s} & 252.4~\si{mm/s} &103.9~\si{mm/s} & 263.3~\si{mm/s}\\ \hline
  \end{tabular}
\end{table}

\begin{table}[t]
  \caption{Experimental results of the $\alpha$ and comsumpted energy $E$ in the case that left and right legs are antiphase}
  \label{tab:performance_anti}
  \centering
  \begin{tabular}{c|c|c|c|c}\hline
    \multirow{2}{*}{Type}&\multicolumn{2}{c|}{Land}&\multicolumn{2}{c}{Water surface}\\\cline{2-5}
     &$\alpha$  &$E$ &$\alpha$ &$E$\\\hline \hline
     Normal&41.4$\pm$2.5~\% &71.0$\pm$3.6~J &44.2$\pm$4.1~\%&44.4$\pm$3.1~J \\
     Fin& 52.6$\pm$5.4~\% &64.8$\pm$6.0~J& 45.6$\pm$0.7~\%&43.2$\pm$6.0~J \\
     Web& 30.7$\pm$1.3~\%& 73.0$\pm$3.5~J& 33.4$\pm$3.6~\%&53.1$\pm$5.6~J \\ \hline
  \end{tabular}
\end{table}

\begin{table}[t]
  \caption{Experimental results of the $\alpha$ and comsumpted energy $E$ in the case that left and right legs are in-phase}
  \label{tab:performance_in}
  \centering
  \begin{tabular}{c|c|c|c|c}\hline
    \multirow{2}{*}{Type}&\multicolumn{2}{c|}{Land}&\multicolumn{2}{c}{Water surface}\\\cline{2-5}
     &$\alpha$ &$E$ &$\alpha$ &$E$\\\hline \hline
     Normal&35.2$\pm$5.4~\% &97.8$\pm$16.6~J &44.5$\pm$4.4~\% &48.3$\pm$11.1~J \\
     Fin& 40.5$\pm$4.6~\% &91.7$\pm$9.3~J& 42.6$\pm$2.7~\% &36.9$\pm$2.3~J \\
     Web& 32.8$\pm$2.9~\% & 84.2$\pm$6.2~J& 39.5$\pm$1.6~\% &43.5$\pm$2.3~J \\ \hline
  \end{tabular}
\end{table}

\begin{figure}[t]
  \begin{minipage}[b]{0.5\columnwidth}
   \centering
   \includegraphics[width=\columnwidth]{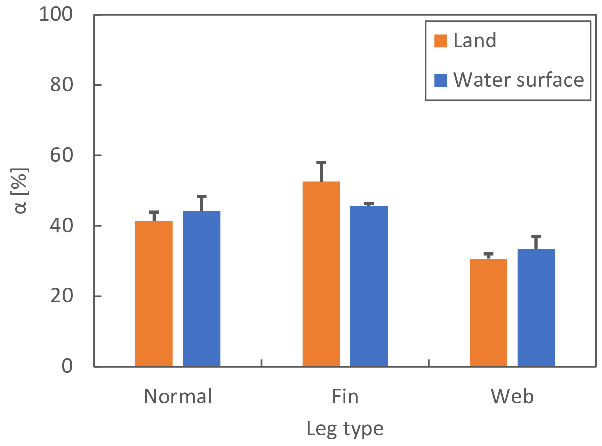}
   \subcaption{Left and right legs are antiphase.}
   \label{fig:alpha_anti}
  \end{minipage}
  \begin{minipage}[b]{0.5\columnwidth}
   \centering
   \includegraphics[width=\columnwidth]{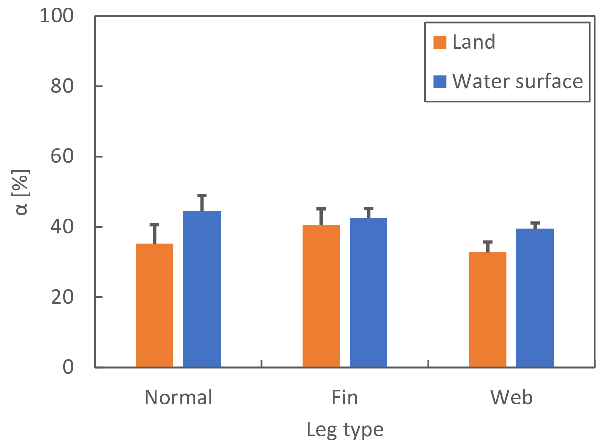}
   \subcaption{Left and right legs are in-phase.}
   \label{fig:alpha_in}
  \end{minipage}
  \caption{Comparison of mobility performance}
  \label{fig:performance}
\end{figure}

\begin{figure}[t]
  \begin{minipage}[b]{0.5\columnwidth}
    \centering
    \includegraphics[width=\columnwidth]{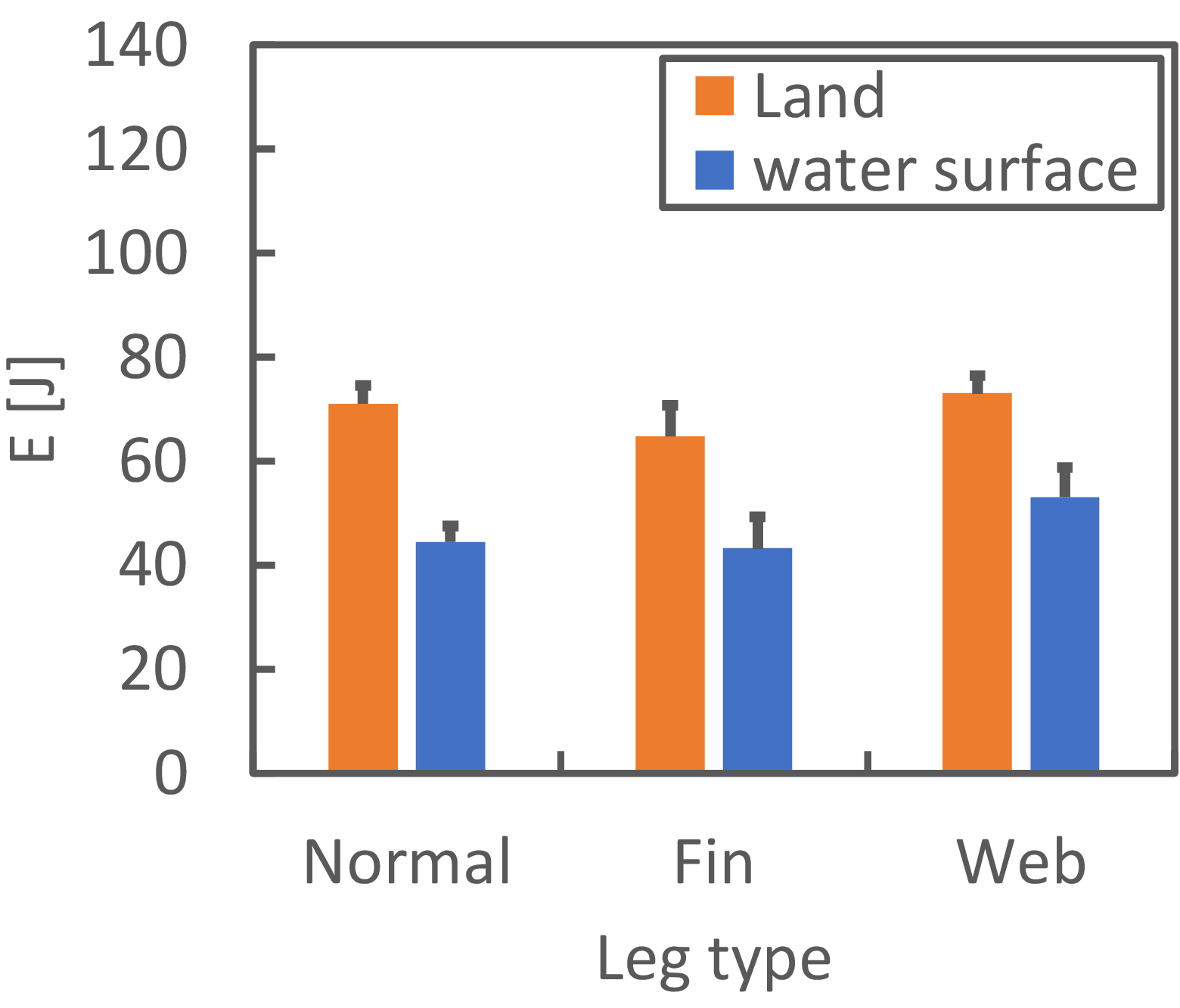}
    \subcaption{Left and right legs are antiphase.}
    \label{fig:E_anti}
  \end{minipage}
  \begin{minipage}[b]{0.5\columnwidth}
    \centering
    \includegraphics[width=\columnwidth]{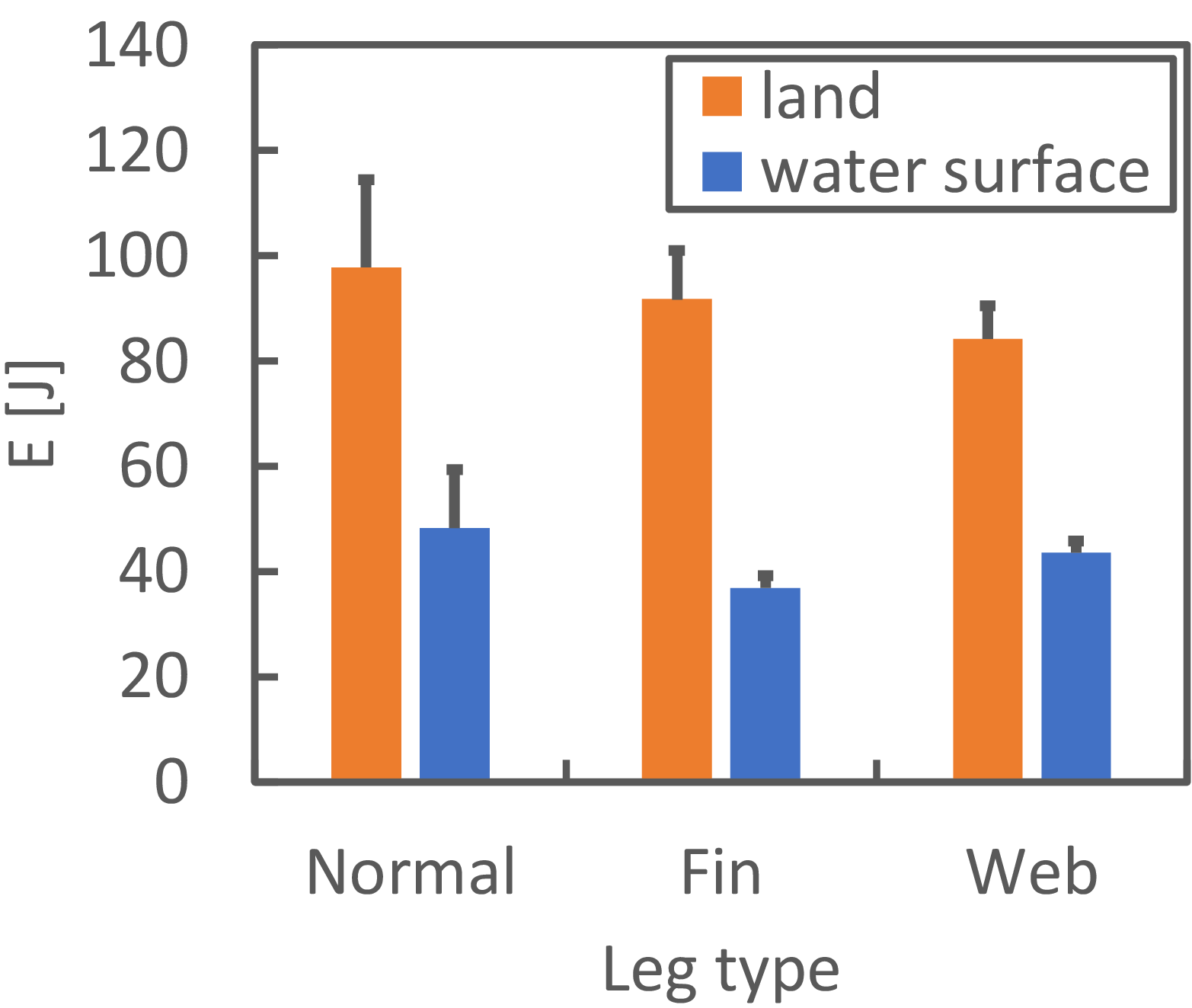}
    \subcaption{Left and right legs are in-phase}
    \label{fig:E_same}
  \end{minipage}
  \caption{Comparison of energy consumption of actuators}
  \label{fig:energy}
\end{figure}

\begin{figure}[t]
  \begin{minipage}[t]{0.5\columnwidth}
   \centering
   \includegraphics[width=\columnwidth]{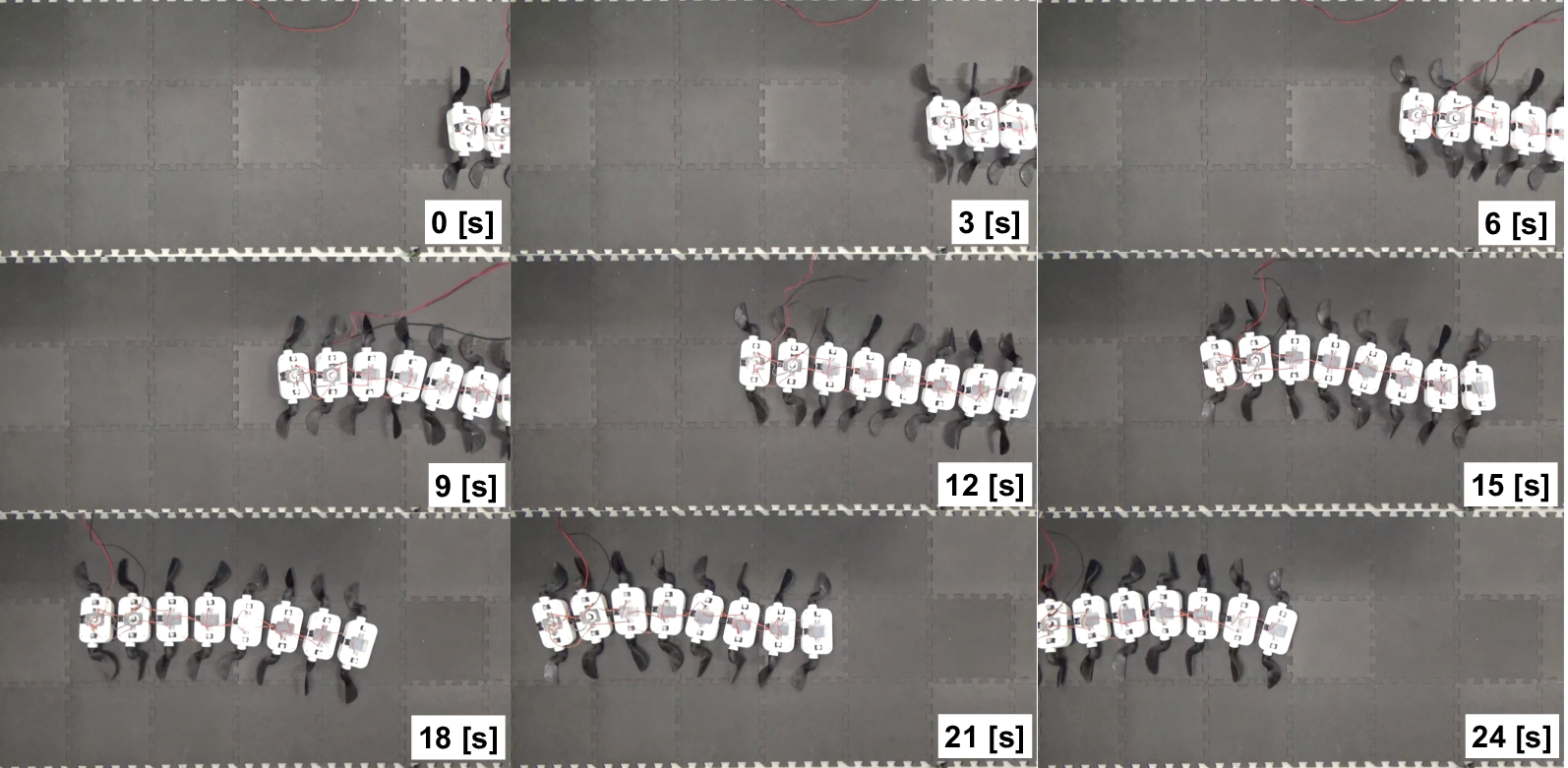}
   \subcaption{Land environment}
   \label{fig:snap_land}
  \end{minipage}
  \begin{minipage}[t]{0.5\columnwidth}
   \centering
   \includegraphics[width=\columnwidth]{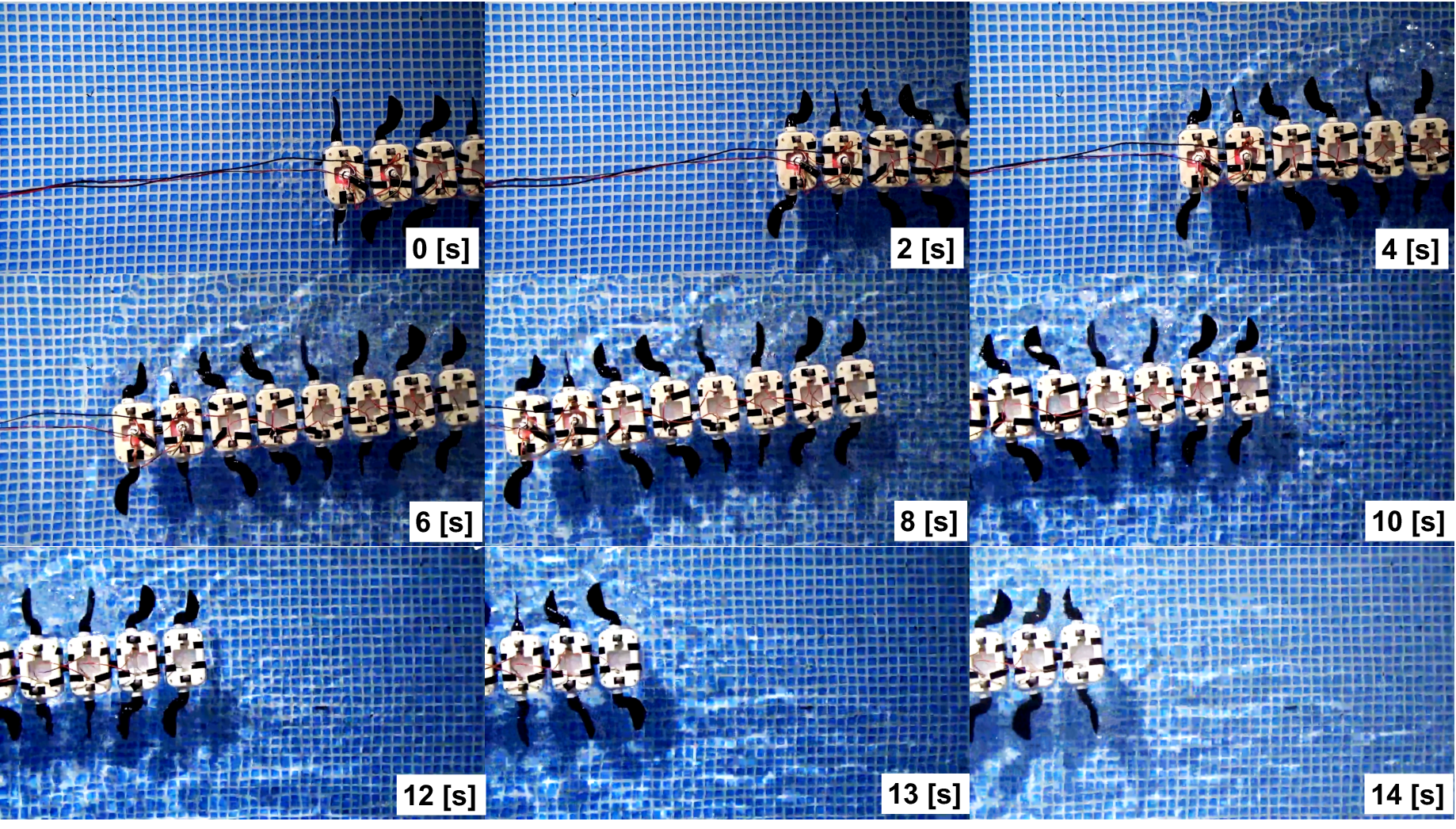}
   \subcaption{Water surface environment}
   \label{fig:snap_water}
  \end{minipage}
  \caption{Snapshots of the movements of the robot when left and right legs are antiphase}
  \label{fig:snap}
\end{figure}

\begin{figure}[t]
  \begin{minipage}[t]{0.5\columnwidth}
    \centering
    \includegraphics[width=\columnwidth]{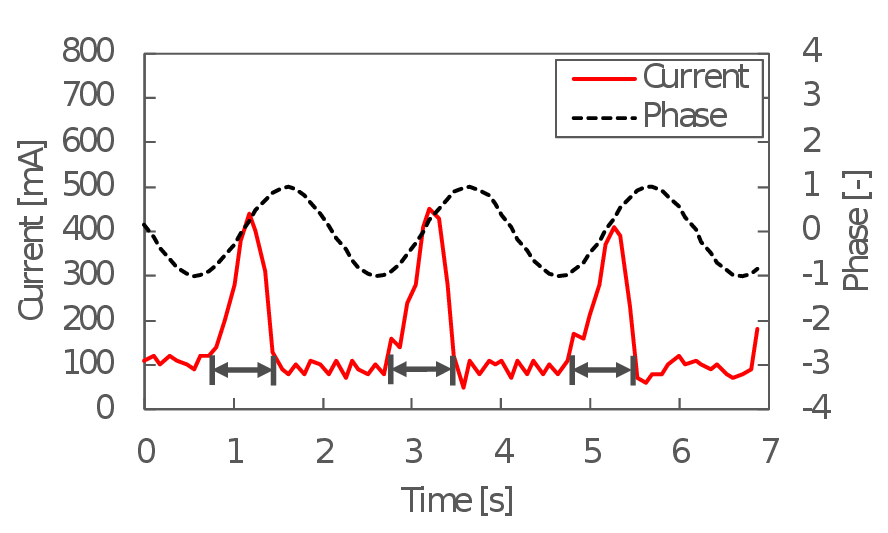}
    \subcaption{Current and phase of Leg~5's motor with Normal type leg}
    \label{fig:currentNormal}
  \end{minipage}
  \begin{minipage}[t]{0.5\columnwidth}
    \centering
    \includegraphics[width=\columnwidth]{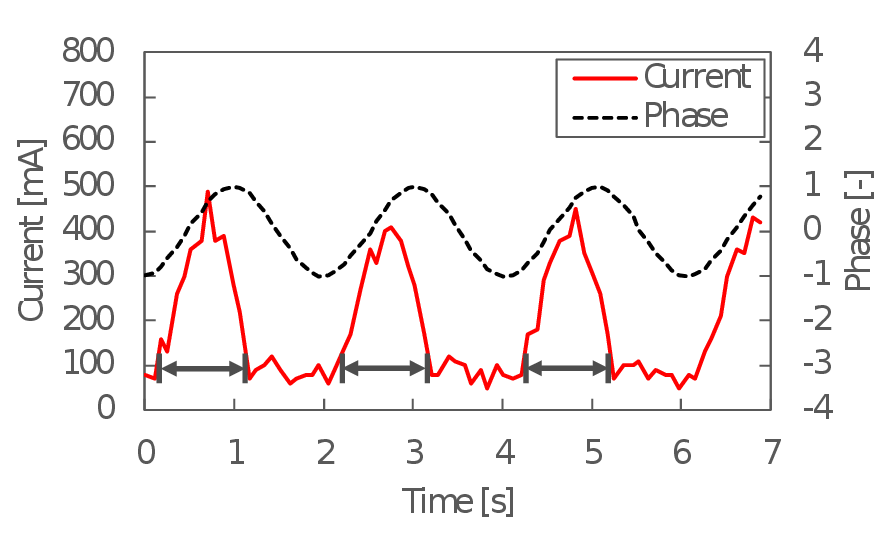}
    \subcaption{Current and phase of Leg~5's motor with Fin type leg}
    \label{fig:currentFin}
  \end{minipage}
  \begin{minipage}[t]{0.5\columnwidth}
    \centering
    \includegraphics[width=\columnwidth]{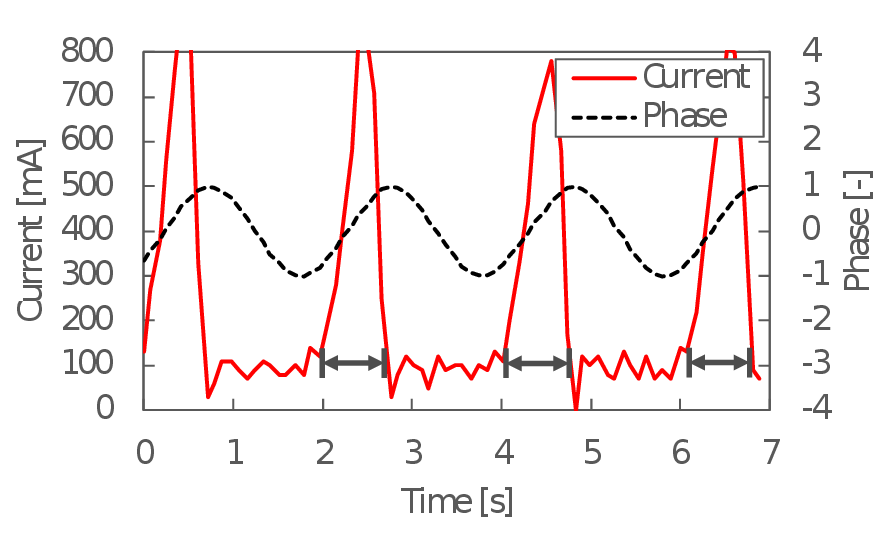}
    \subcaption{Current and phase of Leg~5's motor with Web type leg}
    \label{fig:currentWeb}
  \end{minipage}
  \caption{Current and phase of Leg~5's motor on land when left and right legs are antiphase}
  \label{fig:current}
\end{figure}
\begin{figure}[t]
  \centering
  \includegraphics[width=0.9\columnwidth]{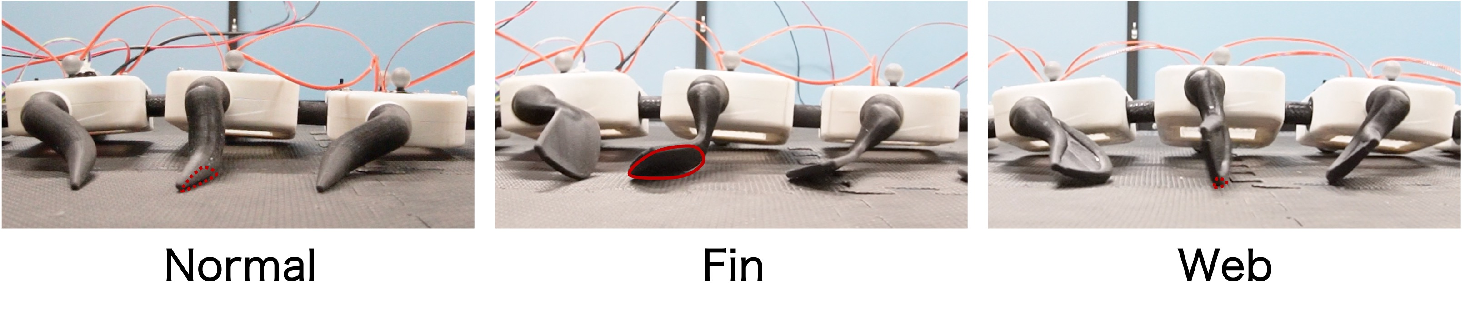}
  \caption{The leg of Fin type is the most grounded of the three types of leg.}
  \label{fig:softfin}
\end{figure}
\begin{figure}[t]
  \begin{minipage}[t]{0.5\columnwidth}
    \centering
    \includegraphics[width=\columnwidth]{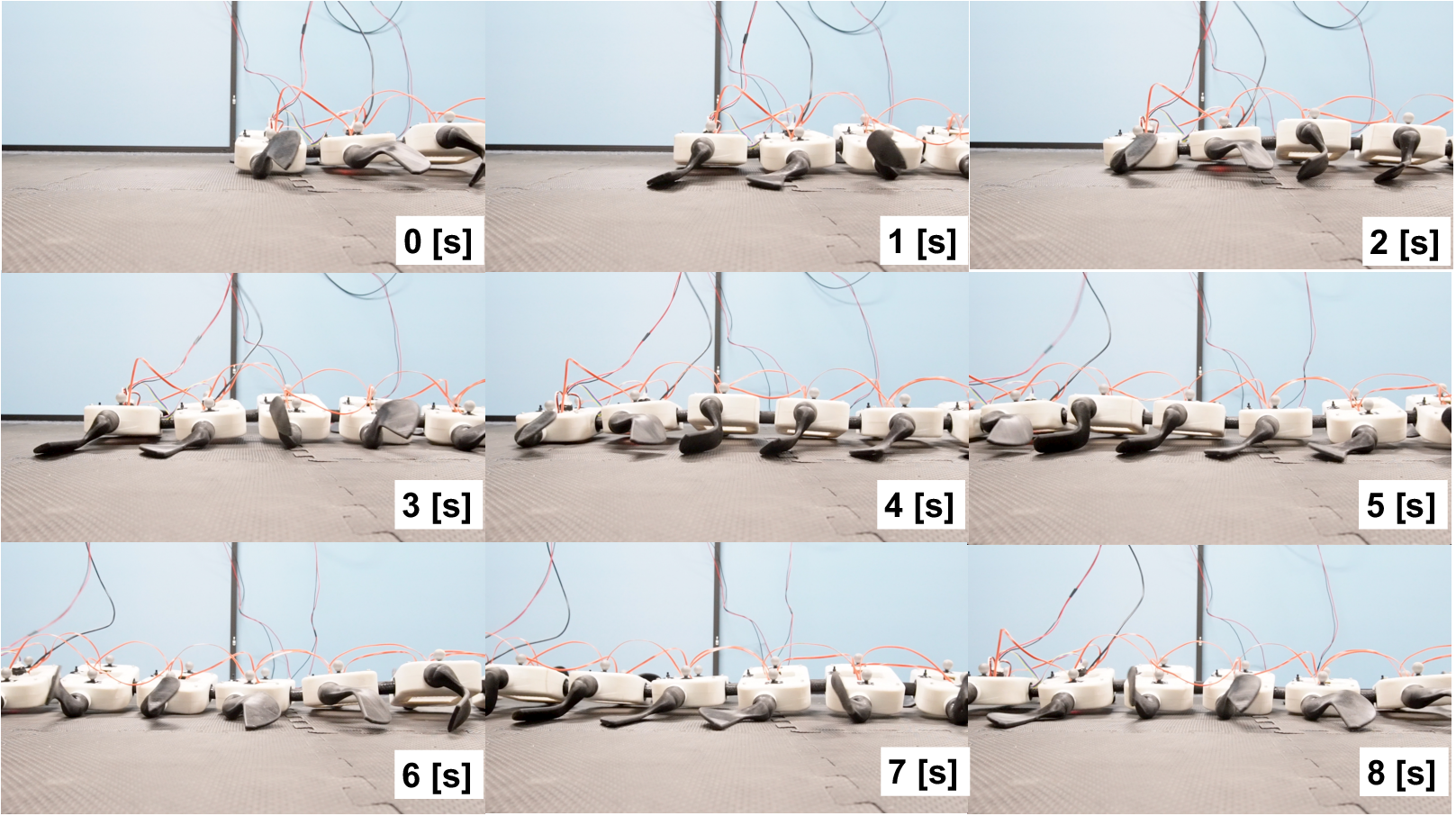}
    \subcaption{Each body segment periodically oscillates in the roll direction when left and right legs are antiphase.}
    \label{fig:wave_anti}
  \end{minipage}
  \begin{minipage}[t]{0.5\columnwidth}
    \centering
    \includegraphics[width=\columnwidth]{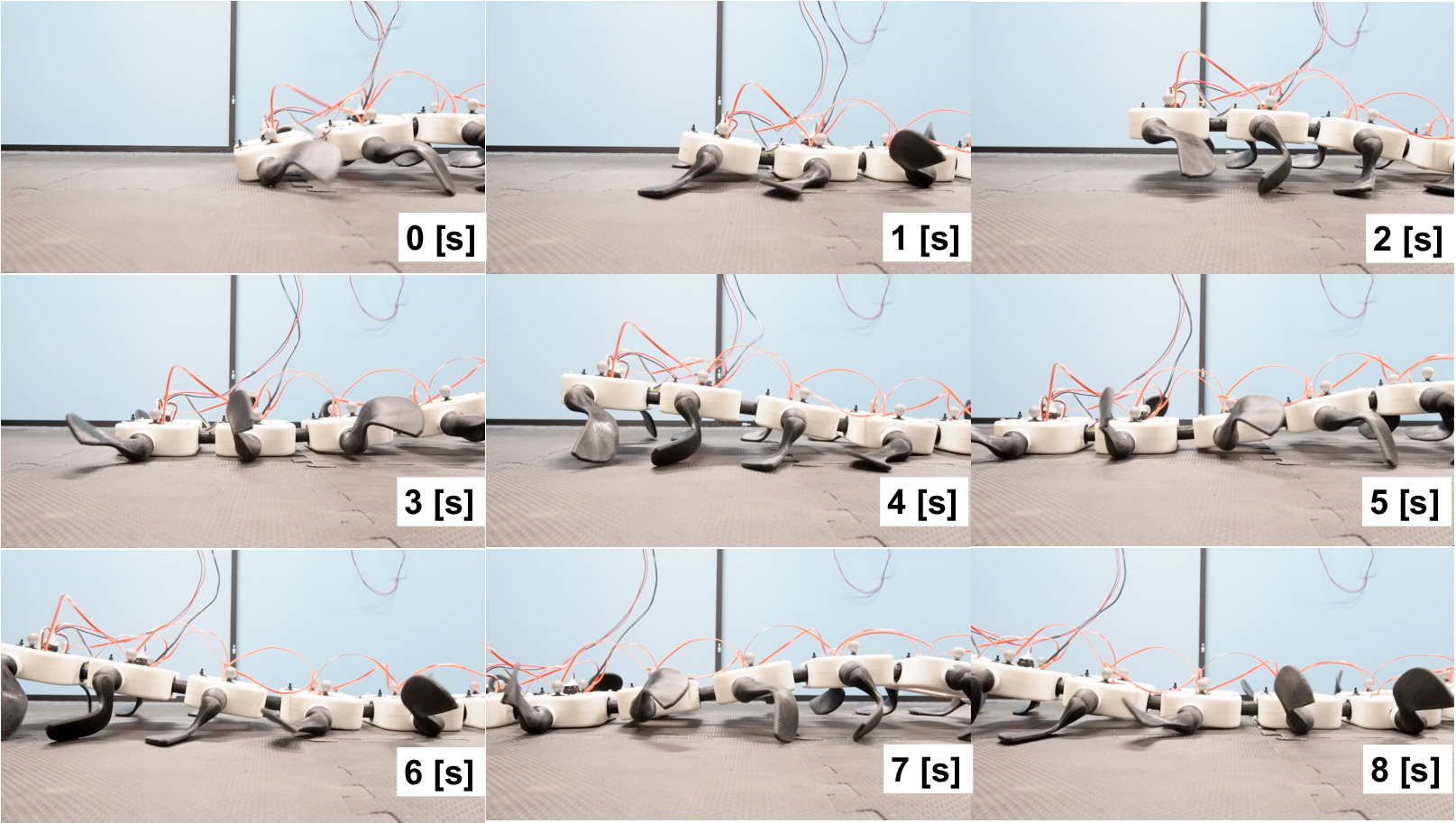}
    \subcaption{The body waves up and down when left and right legs are in-phase.}
    \label{fig:wave_in}
  \end{minipage}
  \caption{The body waves in each phase case of the left and right legs}
  \label{fig:wave}
\end{figure}
\begin{figure}[t]
  \begin{minipage}[t]{0.5\columnwidth}
    \centering
    \includegraphics[width=\columnwidth]{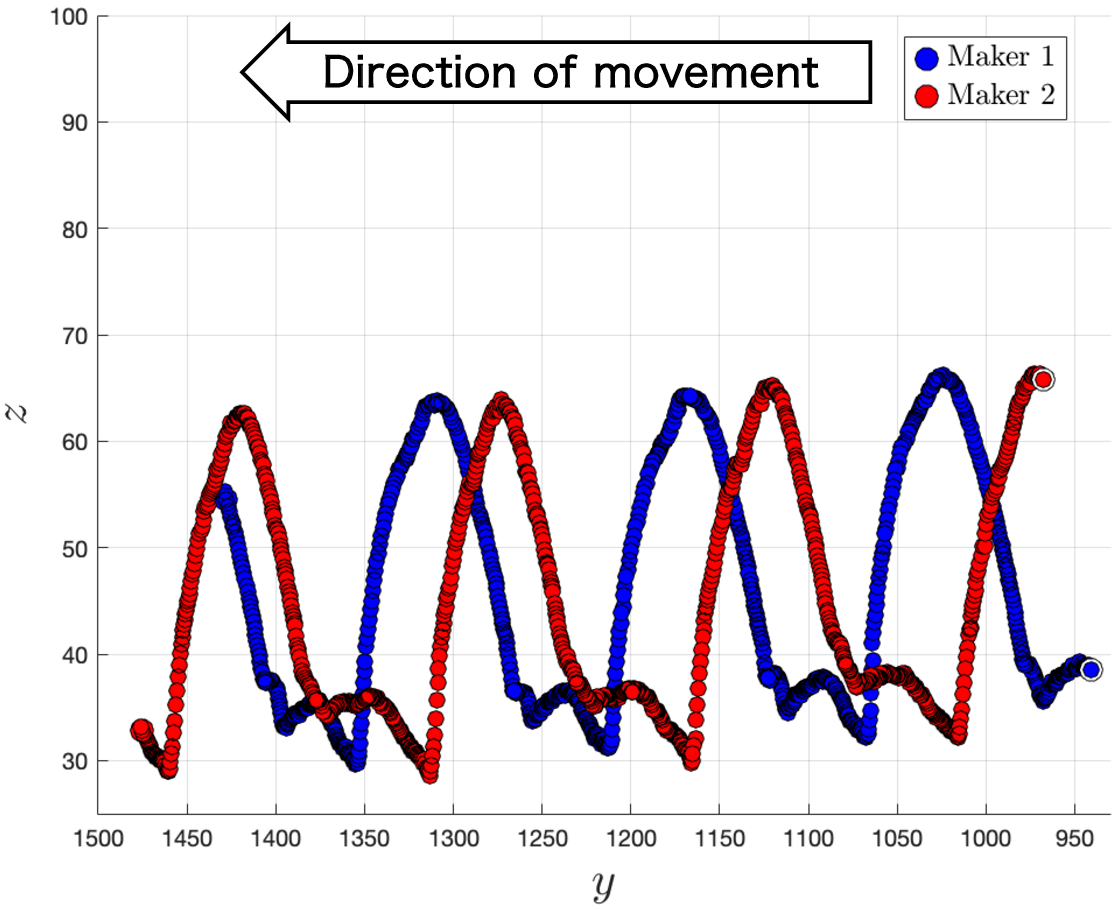}
    \subcaption{Trajectories in the case that left and right legs are antiphase.}
    \label{fig:wave_anti_mocap}
  \end{minipage}
  \begin{minipage}[t]{0.5\columnwidth}
    \centering
    \includegraphics[width=\columnwidth]{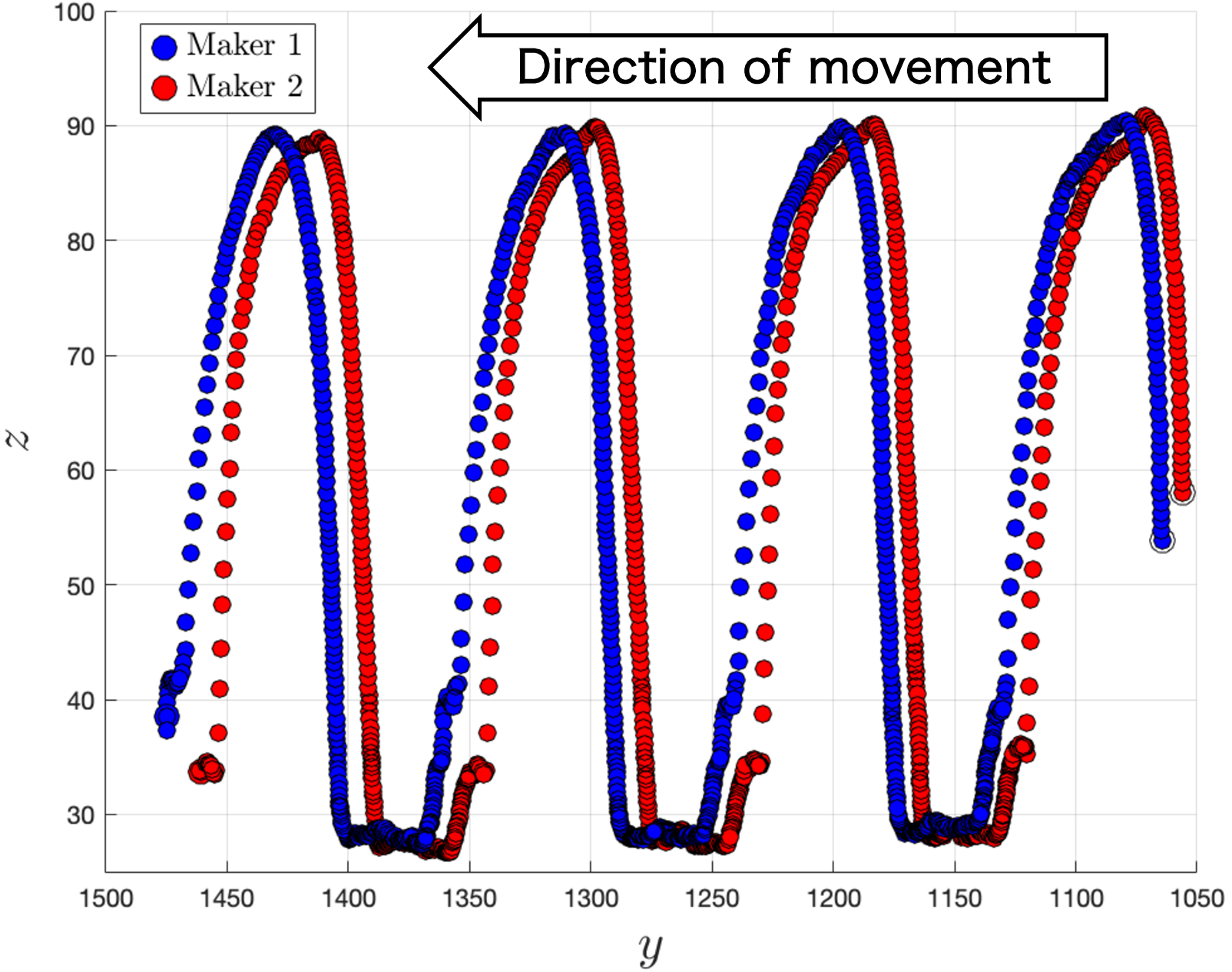}
    \subcaption{Trajectories in the case that left and right legs are in-phase.}
    \label{fig:wave_in_mocap}
  \end{minipage}
  \caption{Trajectory of motion capture markers 1 and 2 on the $yz$ plane}
  \label{fig:wave_mocap}
\end{figure}

\section{Experimental results}
\label{sec:result}
This section evaluates the mobility performance of i-CPA based on the evaluation metrics described in Sec.~\ref{sec:evaluate}.
Table~\ref{tab:performance_anti} and Table~\ref{tab:performance_in} show the average walking speed and average leg circumferential speed of i-CPA.
Table~\ref{tab:performance_anti} shows results for opposite-phase left-right leg phases, while Table~\ref{tab:performance_in} shows results for in-phase left-right leg phases.
Additionally, Table~\ref{tab:performance_anti} and Table~\ref{tab:performance_in} present experimental results for the slip ratio $\alpha$ and the actuator's energy consumption $E$.
Table~\ref{tab:performance_anti} shows results for opposite-phase left-right leg phases, while Table~\ref{tab:performance_in} shows results for in-phase left-right leg phases.
The values for $\alpha$ and $E$ in the tables represent the average and standard deviation from five experiments.
These tables reveal that the relationship between the magnitude of the average walking speed $V$ and that of $\alpha$ differs.
This occurs because, as stated in Section \ref{sec:evaluate}, $V$ depends on the leg rotation radius $r$.

Next, we evaluate mobility performance using $\alpha$.
A bar graph comparing the values of $\alpha$ is shown in Fig.~\ref{fig:performance}.
Hereafter, the five-run average values of $\alpha$ for the Normal, Fin, and Web leg types are denoted as $\alpha_\mathrm{Normal}$~[\%], $\alpha_\mathrm{Fin}$~[\%], and $\alpha_\mathrm{Web}$~[\%], respectively.
When the left and right leg phases were out of phase, the magnitude of $\alpha$ values in both aquatic and terrestrial environments followed the order $\alpha_\mathrm{Fin} > \alpha_\mathrm{Normal} > \alpha_\mathrm{Web}$.
Furthermore, when the left and right leg phases were in phase, the magnitude of $\alpha$ was $\alpha_\mathrm{Fin} > \alpha_\mathrm{Normal} > \alpha_\mathrm{Web}$ on land and $\alpha_\mathrm{Normal} > \alpha_\mathrm{Fin} > \alpha_\mathrm{Web}$ in water.

Furthermore, the results of comparing energy consumption are shown in Fig.~\ref{fig:energy}.
The five-time average values of $E$ for the Normal, Fin, and Web leg types are denoted as $E_\mathrm{Normal}$~[J], $E_\mathrm{Fin}$~[J], and $E_\mathrm{Web}$~[J], respectively.
When the left and right leg phases were out of phase, the magnitude of $E$ in both aquatic and terrestrial environments followed the order $E_\mathrm{Fin} < E_\mathrm{Normal} < E_\mathrm{Web}$.
Furthermore, when the left and right leg phases were in phase, the magnitude of $E$ on land was $E_\mathrm{Web} < E_\mathrm{Fin} < E_\mathrm{Normal}$, while on water, the magnitude of $E$ was $E_\mathrm{Fin} < E_\mathrm{Web} < E_\mathrm{Normal}$. 
When the left and right leg phases were out of phase, the Fin-type legs exhibited the lowest energy consumption in both land and water environments.
Furthermore, when the left and right leg phases were in phase, the Web-type legs had the lowest energy consumption on land, while the Fin-type legs had the lowest energy consumption in water.
Based on these results, it was determined that the i-CPA exhibited the highest mobility performance with Fin-type legs on land and with Normal-type legs in water.

Overall, in the terrestrial environment, the Fin-type legs exhibited the highest slip rate $\alpha$ ($\alpha_\mathrm{Fin} = 52.6 \pm 5.4$~[\%]) and the lowest energy consumption $E$ ($E_\mathrm{Fin} = 64.8 \pm 6.0$~[J]) when the left and right leg phases were out of phase.
In the aquatic environment, the Fin-type leg configuration with opposite-phase left-right leg movement exhibited the highest slip ratio $\alpha$ ($\alpha_\mathrm{Fin} = 45.6 \pm 0.7$~[\%]), while the Fin-type legs with in-phase left-right leg phases exhibited the lowest energy consumption $E$ ($E_\mathrm{Fin} = 36.9 \pm 2.3$ [J]).
Fig.~\ref{fig:snap} shows snapshots of the i-CPA's movement experiments in both land and water environments, using Fin-type legs with left-right legs operating in opposite phases.
Fig.~\ref{fig:snap} demonstrates that the robot can move smoothly in both land and water environments under identical body and control conditions.
These experiments videos can be viewed in the Supplemental Materials (Movie1.mp4 and Movie2.mp4, URL: \url{https://drive.google.com/drive/folders/1MDNh2XfpqPLMOPpXqJojI8oC8THrwEwE?usp=drive_link}).

\section{Discussion}
\label{sec:discussion}
Based on the above results, we first examine the terrestrial locomotion performance using the values of slip ratio $\alpha$ and energy consumption $E$.

First, we examine the terrestrial locomotion performance.
For terrestrial locomotion, driving the fin-type legs with opposite-phase left and right legs yields the maximum $\alpha$ and the minimum $E$.
This is likely because the propulsive force gained from each leg pushing off the ground efficiently contributed to the overall robot propulsion.
Here, Fig.~\ref{fig:current} shows a graph of the current consumption and phase of the motor driving Leg~5 during a 7-second interval of the land experiment under the condition of opposite-phase left and right legs.
The vertical axis represents current consumption and phase, while the horizontal axis represents time.
The red line represents the current consumption, while the black dashed line represents the sine value of the rotation angle from an arbitrary starting point, which corresponds to the phase.
Note that the voltage values for each motor remained nearly constant.
As shown in Fig.~\ref{fig:current}, the current consumption value exhibits periodic changes.
Since the motor is controlled to maintain rotational speed, encountering resistance during rotation increases the current consumption to sustain that speed.
Therefore, periods of periodically increased current consumption correspond to times when the leg encounters resistance from the ground.
Comparing the duration of the peak current consumption—the length of the mountain-like portion in the graph (represented by the length of the double-ended arrow lines in the figure)—for the three leg types reveals that the Fin-type leg exhibits the longest duration.
This trend was similarly observed for the other motors, indicating that the Fin-type leg experiences resistance from the ground for the longest period.
Fig.~\ref{fig:softfin} illustrates the ground contact state for the three leg types.
As shown in Fig.~\ref{fig:softfin}, the Fin-type leg experienced resistance from the ground for the longest duration.
This is because the Fin-type leg is the softest and has the largest contact area.
Therefore, the Fin-type leg exhibited the highest mobility performance because, among the three legs, it experienced resistance from the ground for the longest duration and efficiently obtained propulsive force relative to the ground.

Conversely, when the left and right leg phases are in phase, mobility performance on land deteriorates compared to when they are out of phase.
We will examine the cause of this.
Fig.~\ref{fig:wave} shows the wave-like motion of the i-CPA body during experiments when the left and right leg phases are in phase and out of phase.
Additionally, Fig.~\ref{fig:wave_mocap} shows the movement trajectories of motion capture markers 1 and 2 attached to the i-CPA body in the $yz$ plane (as seen from the robot's side).
The blue dotted line indicates the trajectory of marker 1 shown in Fig.~\ref{fig:number}, while the red dotted line indicates the trajectory of marker 2.
As shown in Fig.~\ref{fig:wave} and Fig.~\ref{fig:wave_mocap}, when the left and right leg phases are out of phase, each segment of the i-CPA periodically oscillates in the roll direction. Conversely, when the left and right leg phases are in phase, the body undulates vertically.
This confirms that when the left and right legs are in phase, the force exerted by the legs pushing against the ground acts primarily in the vertical upward direction, rather than the horizontal direction, compared to when they are out of phase.
In other words, in this case, it is considered that the force exerted by the legs pushing against the ground did not contribute significantly to propulsion, regardless of the type of leg.
These experiments videos can be viewed in the Supplemental Materials (Movie3.mp4 and Movie4.mp4, URL: \url{https://drive.google.com/drive/folders/1MDNh2XfpqPLMOPpXqJojI8oC8THrwEwE?usp=drive_link})
Moving forward, we will attach pressure sensors to the legs to determine the direction of the reaction force exerted on the ground by the legs. This will enable us to quantitatively evaluate the propulsive force applied to the ground and conduct a more detailed examination of the factors contributing to enhanced mobility performance on land.

Next, we examine mobility performance on water.
For water mobility, driving the Fin-type legs with opposite phase between left and right legs resulted in the highest $\alpha$.
Regarding energy consumption $E$, driving the fin-type legs in phase yielded the lowest value.
However, when the legs were out of phase, there was little difference in $\alpha$ between the fin-type and normal-type legs, while the webbed-type leg showed a lower $\alpha$ compared to them.
Furthermore, when the legs were in phase, no significant difference in mobility performance was observed across all leg shapes.
This indicates that while the Fin and Web leg designs aimed to increase thrust by enlarging the area of the water-propelling section, this effect was minimal on land.
Based on the above analysis, among the three leg types, the Fin-type leg is considered the most suitable for locomotion in both terrestrial and aquatic environments. Specifically, it exhibited the highest locomotion performance on land and comparable performance to the Normal-type leg in water when the left and right legs were out of phase.

The results above indicate that legs with soft, large contact areas are advantageous for terrestrial locomotion.
Conversely, regarding legs suited for aquatic locomotion, it was suggested that larger paddle surface areas do not necessarily enhance locomotive performance.
However, differences in the relative performance of leg types were observed in both terrestrial and aquatic environments depending on whether the left and right leg phases were in opposite or in phase. This indicates that changing the phase difference between the front and rear legs may alter the relative performance of leg types.
In particular, the phase difference between the legs appears to be more critical for aquatic locomotion.
Long-tailed crustaceans like krill swim by rhythmically moving their swimming appendages.
Reference \cite{metachronal} describes a movement called metachronal swimming, where adjacent swimming appendages maintain a phase difference of approximately 1/4 cycle. Numerical fluid dynamics modeling revealed this movement to be the most energy-efficient for these crustaceans.
Thus, it is suggested that an ideal phase difference between front and rear legs exists in the body structure of organisms in nature.
Drawing on this finding, future research will conduct experiments with various front-to-back leg phase differences to investigate which leg configurations are best suited for locomotion in both terrestrial and aquatic environments.
Additionally, factors like leg rotation speed and joint flexibility may also influence the relative performance of different leg types.
Considering these various elements should enable a discussion on the optimal leg shape for efficient locomotion across both terrestrial and aquatic environments.

\section{Conclusion}
\label{sec:conc}
This study aimed to further develop mobile robots based on implicit control, targeting the creation of a flexible multi-legged robot capable of moving freely in both land and water environments using a single, simple explicit control scheme independent of the environment. Consequently, the amphibious flexible multi-legged robot i-CentiPot-Amphibian was developed, featuring flexible legs and joints in an 8-segment, 16-legged configuration.
Furthermore, recognizing that the legs are the primary point of contact with the environment and that optimizing leg shape is key to leveraging environmental interactions, we focused on leg design, creating three leg types: Normal, Fin, and Web.
We conducted evaluation experiments on the locomotion performance of these legs on land and water using a single positive control regardless of the environment to investigate which leg shape is most suitable for locomotion in both terrestrial and aquatic environments.
The results showed that the Fin-type leg, with its soft, fin-like shape, was most suitable for locomotion in both terrestrial and aquatic environments.
Specifically, it was found that a soft leg with a large contact area is likely advantageous for terrestrial locomotion.
Conversely, regarding legs suitable for aquatic locomotion, it was suggested that a larger area for scooping water does not necessarily guarantee higher locomotion performance.
Moving forward, to further develop the discussion on leg shapes suitable for movement in mixed terrestrial and aquatic environments, we plan to evaluate locomotion performance by varying various parameters such as phase differences between legs, flexibility of body joints, and leg rotation speed.

\section*{Acknowledgements}
This research was supported in part by grants-in-aid for JSPS KAKENHI Grant Number JP25K17629 and JST Moonshot Research and Development Program JPMJMS2032 (Innovation in Construction of Infrastructure with Cooperative AI and Multi Robots Adapting to Various Environments).

\bibliographystyle{tfnlm}
\bibliography{myref_multileg}

\end{document}